%% file: main.tex
\documentclass[acmtog,screen]{acmart} 
\acmSubmissionID{745}

\copyrightyear{2025}
\acmYear{2025} 
\acmConference[SIGGRAPH Conference Papers '25]{}{August 10--14, 2025}{Vancouver, BC, Canada}
\acmDOI{10.1145/3721238.3730691}
\acmISBN{979-8-4007-1540-2/2025/08}

\input{config/packages}
\input{config/formats}

\input{config/pronoun}
\begin{document}

\title{\ourtitle} 
\input{config/authors}

\input{sec/0_abstract}

\include{config/ccs_concepts}

\input{figs/teaser}

\maketitle

\input{sec/1_intro} 
\input{sec/2_related}
\input{sec/4_method_paul}

\input{sec/5_experiments}

\input{sec/6_conclusion}
\input{sec/acknowledgement}

\bibliographystyle{ACM-Reference-Format}
\bibliography{reference}

\clearpage 
\input{figs/combine_2_figures}
\input{figs/diverse_avatars}

\input{figs/comparison}

\clearpage
\twocolumn[%
\begin{@twocolumnfalse}
\vspace{2em}
\begin{center}
{\fontsize{22pt}{28pt}\selectfont Supplementary Material}
\end{center}
\vspace{1.5em}
\end{@twocolumnfalse}
]

\input{sec/appendix}

\end{document}

%% file: config/packages.tex
\usepackage{graphicx} 
\usepackage[capitalize]{cleveref}
\usepackage{booktabs}
\usepackage{orcidlink}

\usepackage{wrapfig}
\usepackage{multirow}
\usepackage{tabularx}
\usepackage{algorithmic}
\usepackage[ruled]{algorithm2e}

\SetAlFnt{\small}
\SetAlCapFnt{\small}
\SetAlCapNameFnt{\small}
\SetAlCapHSkip{0pt}

\usepackage{mathtools}
\usepackage{balance}

\usepackage{float}
\usepackage{enumitem}
\usepackage{comment}
\usepackage{pifont}
\usepackage[toc,page]{appendix}

\usepackage{soul}
\usepackage{acronym}
\usepackage{fancybox}
\usepackage{epigraph}
\usepackage{dirtytalk}
\usepackage{bm}
\usepackage{fontawesome}
\usepackage{colortbl} 

\usepackage{xr}
\usepackage{tikz} 
\usepackage{transparent}
\usepackage{subfig}
\usepackage{hyphenat} 
\usepackage{microtype}

%% file: config/formats.tex
\definecolor{cvprblue}{rgb}{0.21,0.49,0.74}
\definecolor{citecolor}{HTML}{0071bc}
\definecolor{frontcolor}{HTML}{325ea5}
\definecolor{backcolor}{HTML}{a58b77}
\definecolor{sidecolor}{HTML}{10768c}
\definecolor{skincolor}{HTML}{dcb7b7}
\definecolor{darkred}{rgb}{0.6, 0.1, 0.05}
\definecolor{DeltaColor}{rgb}{0.039,0.73,0.71}
\definecolor{SigmaColor}{rgb}{0.98,0.45,0.0}
\definecolor{AlphaColor}{rgb}{0,0,0.8}
\definecolor{BetaColor}{rgb}{0.8,0,0.8}
\definecolor{GammaColor}{rgb}{0.514,0.34,0.224}
\definecolor{EpsilonColor}{rgb}{0.353,0.725,0.906}
\definecolor{PurpleColor}{HTML}{9839ff}
\definecolor{BadColor}{HTML}{C0392B}
\definecolor{OrangeColor}{rgb}{0.914,0.541,0.0.141}
\definecolor{GreenColor}{HTML}{00ab41}
\definecolor{LightBlue}{HTML}{7dbaf3}
\definecolor{RedColor}{rgb}{0.949,0.275, 0.224}
\definecolor{LightCyan}{rgb}{0.88,1,1}
\definecolor{Gray}{gray}{0.85}
\definecolor{LightGray}{gray}{0.70}
\definecolor{PinkColor}{HTML}{f37dba}

\definecolor{greenprior}{HTML}{34a853}
\definecolor{redprior}{HTML}{ea4335}
\definecolor{blueprior}{HTML}{4285f4}

\definecolor{bestcolor}{rgb}{1, 0.5, 0.25}
\definecolor{secondbestcolor}{rgb}{1, 0.8, 0.5}

\newcolumntype{a}{>{\columncolor{Gray}}c}

\newcommand{\qheading}[1]{\noindent\textbf{#1.}}

\makeatletter
\newcommand*{\addFileDependency}[1]{%
  \typeout{(#1)}
  \@addtofilelist{#1}
  \IfFileExists{#1}{}{\typeout{No file #1.}}
}
\makeatother

\newlength\savewidth

%% file: config/pronoun.tex
\newcommand{\page}{\href{https://github.com/TingtingLiao/soap}{\textcolor{magenta}{\xspace\tt\textit{github.com/TingtingLiao/soap}}\xspace}}

\newcommand{\norm}[1]{\left\lVert#1\right\rVert}

\newcommand{\nerf}{\mbox{NeRF}\xspace}

\newcommand{\flame}{\mbox{FLAME}\xspace}    

\newcommand{\modelname}{\mbox{SOAP}\xspace} 
\newcommand{\modelnameLong}{Style-Omniscient Animatable Portraits}  
\newcommand{\ourtitle}{\modelname: \modelnameLong}

\newcommand{\rebuttal}[1]{\textcolor{black}{#1}}

%% file: config/authors.tex
\author{Tingting Liao}
\email{tingting.liao@mbzuai.ac.ae}
\orcid{0000-0003-1310-3810}
\affiliation{%
  \institution{Mohamed bin Zayed University of Artificial Intelligence}
  \country{UAE}
} 

\author{Yujian Zheng}
\email{yujian.zheng@mbzuai.ac.ae}
\orcid{0000-0001-7784-8323}
\affiliation{%
  \institution{Mohamed bin Zayed University of Artificial Intelligence}
  \country{UAE}
} 

\author{Adilbek Karmanov}
\email{adilbek.karmanov@mbzuai.ac.ae}
\orcid{0009-0008-4923-5309}
\affiliation{%
  \institution{Mohamed bin Zayed University of Artificial Intelligence}
  \country{UAE}
}

\author{Liwen Hu}
\email{liwen@pinscreen.com}
\orcid{0000-0002-6614-5785}
\affiliation{%
  \institution{Pinscreen}
  \country{USA}
} 

\author{Leyang Jin}
\email{leyang.jin@mbzuai.ac.ae}
\orcid{0009-0002-1440-9096}
\affiliation{%
  \institution{Mohamed bin Zayed University of Artificial Intelligence}
  \country{UAE}
} 

\author{Yuliang Xiu}
\email{xiuyuliang@westlake.edu.cn}
\orcid{0000-0003-0165-5909}
\affiliation{%
  \institution{Westlake University}
  \country{China}
}

\author{Hao Li}
\email{hao@hao-li.com}
\orcid{0000-0002-4019-3420}
\affiliation{%
  \institution{Mohamed bin Zayed University of Artificial Intelligence}
  \country{UAE}
} 
\affiliation{%
  \institution{Pinscreen}
  \country{USA}
}

%% file: sec/0_abstract.tex
\begin{abstract}


Creating animatable 3D avatars from a single image remains challenging due to style limitations (realistic, cartoon, anime) and difficulties in handling accessories or hairstyles.
While 3D diffusion models advance single-view reconstruction for general objects, outputs often lack animation controls or suffer from artifacts because of the domain gap.
We propose SOAP, a style-omniscient framework to generate rigged, topology-consistent avatars from any portrait. 
Our method leverages a multiview diffusion model trained on 24K 3D heads with multiple styles and an adaptive optimization pipeline to deform the FLAME mesh while maintaining topology and rigging via differentiable rendering.
The resulting textured avatars support FACS-based animation, integrate with eyeballs and teeth, and preserve details like braided hair or accessories. 
Extensive experiments demonstrate the superiority of our method over state-of-the-art techniques for both single-view head modeling and diffusion-based generation of Image-to-3D. 
Our code and data are publicly available for research purposes at \page. 

\end{abstract}

%% file: config/ccs_concepts.tex
\ccsdesc[500]{Computing methodologies~Shape modeling; Neural networks}

%
%

\keywords{head modeling, single-view reconstruction, deep neural networks.}

%% file: figs/teaser.tex
\begin{teaserfigure}
    \centering
    \includegraphics[width=0.98\textwidth]{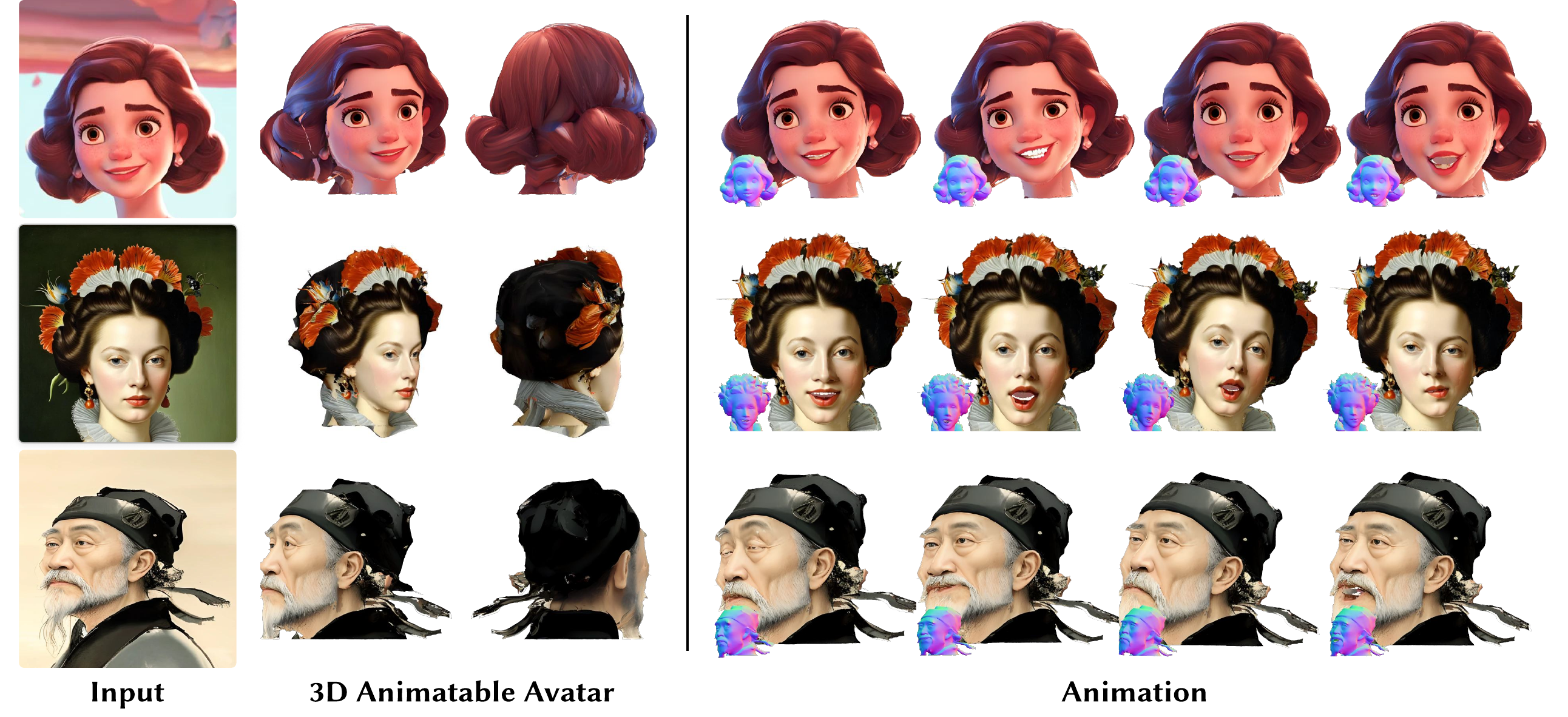}
    \caption{
       \modelname can reconstruct well-rigged 3D heads with eyeballs and teeth, from a single image across various styles. 
       The reconstructed models are fully animatable with facial expressions, natural eye movements, and lifelike lip motions. 
    }
    \label{fig:teaser}
\end{teaserfigure}

%% file: sec/1_intro.tex
\section{Introduction}
\label{sec:intro} 

Whether in storytelling or virtual worlds, human characters are not confined to a realistic look; they span a wide spectrum of styles. Beyond the diverse cartoon aesthetics -- such as those found in Disney, Pixar, or Anime -- avatars can also feature unique hairstyles and various accessories, from hats to glasses, adding further layers of personality and customization.
The ability to generate fully animatable 3D avatars from just a single input image -- be it a photograph or a drawing -- is especially compelling, as it significantly streamlines character production in games and films. This capability also opens up new possibilities for interactive 3D applications, such as virtual reality and gaming, where creating customized 3D avatars becomes as effortless and accessible as taking a photo.

Current state-of-the-art methods for single-view head modeling are often constrained to specific styles, such as photorealism~\cite{Khakhulin2022ROME} or certain cartoon genres~\cite{chen2023panic}, and frequently encounter challenges with accessories like glasses or headgear. Although recent advancements in 3D diffusion-based techniques have shown impressive one-shot modeling capabilities for general objects~\cite{long2024wonder3d,tang2025lgm,wu2024unique3d}, domain-specific content, such as human faces, often lacks fine detail and is prone to unwanted artifacts. Additionally, the 3D outputs are typically either unstructured surface models or neural fields, which are not directly suitable for facial animation and require a separate fitting process using parametric template models, such as FLAME~\cite{FLAME:SiggraphAsia2017} and 3DMM~\cite{pascal20093dmm}.

We introduce the first full-head reconstruction technique from a single portrait that is truly \textit{style-omniscient}, capable of handling realistic faces as well as a broad spectrum of cartoon styles and hairstyles. Our approach generates a high-quality textured parameterized 3D model with clean mesh topology in the face area, complete with an animation rig, including optimized eyeballs and teeth models, while accurately capturing diverse hairstyles and head accessories. We focus on the generation of textured meshes with FACS-based parametric controls, as these are the most prevalent 3D representations in today’s interactive applications. This choice enables efficient rendering, seamless integration with game engines (such as Unreal and Unity), and provides artist-friendly controls -- unlike radiance-field representations like NeRFs and Gaussian fields.

Our approach starts from generating sparse but high-resolution views (6 images and normal maps) from a single portrait input. To accommodate various styles, hairstyles, and accessories, we fine-tune a generic multi-view diffusion model~\cite{wu2024unique3d} using a large-scale (24K) 3D head dataset. Unlike existing image-to-3D generative models that reconstruct static meshes, we produce well-rigged and animatable outputs. We employ an adaptive remeshing and rig optimization technique grounded in differentiable rendering. This approach gradually forms the target avatar by deforming its vertices, correcting its topology, and updating its skinning weights, beginning with a FLAME model.

Our image-to-avatar pipeline demonstrates strong robustness and generalization capabilities, successfully handling a wide variety of styles — from photorealistic portraits to highly stylized cartoon renderings. It is capable of faithfully reconstructing complex hairstyles and accurately preserving diverse head accessories, including hats, glasses, and jewelry. Our contributions are summarized as follows:

\begin{itemize}[leftmargin=*]
    \item A style-omniscient Image-to-Avatar \textbf{pipeline} that reconstructs a fully textured, topology-consistent, and well-rigged mesh-based avatar (with eyeballs and teeth) from a single portrait image across a wide range of styles, haircuts, and accessories.
    \item A multi-view diffusion \textbf{model}, trained on a comprehensive large-scale (24K) \textbf{dataset} of 3D heads, generates consistent views of human head models in various styles.
    \item A differentiable rendering-based deformation \textbf{technique} with adaptive remeshing and rigging that can register any stylized avatar to a parametric head model while maintaining correct semantic correspondence.
\end{itemize}

%% file: sec/2_related.tex
\section{Related Work}
\label{sec:related}

\input{figs/pipe}

\qheading{Animatable Head Modeling} 
Parametric 3D head models are widely used as statistical priors for animatable head modeling. 
3D Morphable Models (3DMMs) \cite{pascal20093dmm} represent head shapes using low-dimensional principal components. 
Building on this, \flame\cite{FLAME:SiggraphAsia2017} introduces both shape and pose blendshapes, enabling expression and movements of the jaw, neck, and eyeballs.  
Subsequent works \cite{danvevcek2022emoca,yao2021deca, Feng2023DELTA} leverage parametric head models \cite{blanz2023morphable,ploumpis2020towards,FLAME:SiggraphAsia2017} to model detailed expressions and emotions. 
ROME \cite{Khakhulin2022ROME} introduces the vertex offset to capture the hair geometry.  
However, these methods often produce overly smooth surfaces due to fixed topologies and limited representation power, struggling with complex geometries like headwear or intricate hairstyles.
Another line of research explores hybrid representations for 3D head modeling. DELTA \cite{Feng2023DELTA} combines explicit meshes for facial regions with NeRF-based hair modeling, enabling diverse hairstyles.

To achieve high-quality rendering, several works \cite{gafni2021dynamic,grassal2022neural,xu2023avatarmav} adopt neural radiance fields (\nerf) \cite{mildenhall2021nerf} to model head avatars. 
HeadNeRF \cite{hong2022headnerf} introduces a parametric model \nerf that integrates the head model into \nerf, while INSTA \cite{zielonka2023instant} develops a dynamic \nerf based on InstantNGP \cite{mueller2022instant}. PointAvatar \cite{zheng2023pointavatar} presents a point-based representation, learning the deformation field based on FLAME’s expression to control the points.  NeRFBlendshape \cite{gao2022reconstructing} constructs NeRF-based blendshape models, combining multi-level voxel fields with expression coefficients to achieve semantic animation control and photorealistic rendering. 

Recently, there are approaches \cite{ma2024gaussianblendshapes,saito2024codec,chen2024monogaussianavatar,dhamo2025headgas,qian2024gaussianavatars,wang2023gaussianhead} utilizing 3D Gaussian Splatting \cite{kerbl3Dgaussians} to model head avatars. 
FlashAvatar \cite{xiang2024flashavatar} attaches Gaussians on a mesh with learnable offsets. 
GuassianBlendshapes \cite{ma2024gaussianblendshapes} decomposes the offsets to blendshapes. 
Though effective for realistic avatars, these methods struggle with stylized content.

\qheading{Generative Head Modeling}  
Recent advances in head modeling \cite{An2023PanoHead,gu2024diffportrait3d,gu2025diffportrait360,li2024spherehead,wang2023rodin,zhang2024rodinhd} have utilized generative models for novel view synthesize. 
PanoHead \cite{An2023PanoHead} uses a tri-grid neural volume representation, allowing 360-degree head synthesis. 
Rodin \cite{wang2023rodin} and its extension RodinHD \cite{zhang2024rodinhd} adopt the diffusion model to generate a triplane of a human head. 
However, these generated heads are static and not suitable for animation. 
Liveportrait \cite{guo2024liveportrait} animates single images into dynamic videos but operates in 2D space. 
CAT4D \cite{wu2024cat4d} trained a multiview morphable diffusion model to create dynamic avatars. 
However, diffusion-based methods often face challenges with cross-view consistency. Another line of work \cite{chen2023fantasia3d,qian2023magic123,lin2023magic3d,tang2023dreamgaussian,liao2024tada} focuses on distilling 2D diffusion priors into 3D through score distillation sampling (SDS). Although high quality is achieved, these require hours per avatar. In contrast, feedforward methods \cite{hong2023lrm,tang2025lgm,xu2024grm} are able to generate 3D assets within seconds after training on large-scale 3D datasets. However, since these methods are trained with general object datasets, there is a significant domain gap when applied to human heads, often yielding inaccurate head shapes. In general, these inference-based methods remain limited to reconstructing static avatars.


%% file: figs/pipe.tex
\begin{figure*}%
\centering
    \includegraphics[width=0.95\textwidth]{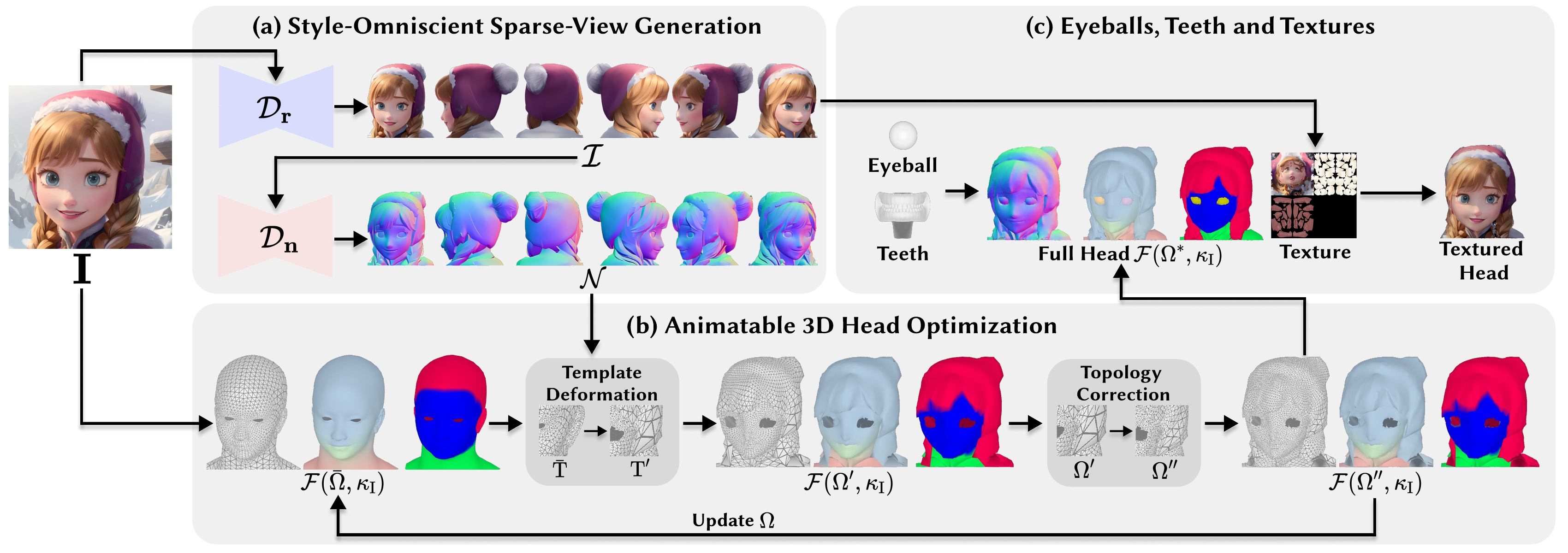}
    \caption{
        Method overview. Given an input image $\mathrm{I}$, \modelname (a) generates six orthogonal RGB images $\mathcal{I}$ and normal images $\mathcal{N}$, then (b) 
        deforms the \flame mesh $\mathcal{F}(\bar{\Omega}, \kappa_{\mathrm{I}})$ to $\mathcal{F}({\Omega}^*, \kappa_{\mathrm{I}})$, 
        and (c) fits eyeballs and teeth to the mesh and generates the texture map.
        }
    \label{fig:pipeline}
\end{figure*}

%% file: sec/4_method_paul.tex

\section{Overview and Preliminary}
Given a 2D portrait, \modelname aims to create a well-rigged and animatable 3D head avatar with detailed geometry and comprehensive texture. However, the diversity in appearance and shape presents significant challenges for reconstructing an animation-ready avatar from style-agnostic portrait images.

Our key insight tackles this challenge in two main aspects. To capture the diverse styles, we harness the power of diffusion models to learn and generalize both appearance and geometry for consistent representation across multiple views. To accommodate varying head shapes, we developed an optimization process that adaptively deforms the initial well-rigged and parameterized shapes to fit different geometries while preserving the semantic features of the head. For example, the original mouth is deformed towards the target mouth rather than the nose.

\qheading{Preliminary}
\label{sec:preliminary}
\flame \cite{FLAME:SiggraphAsia2017} is a parametric human head model. Given the shape $\beta$, pose $\theta$ and expression $\psi$ parameters, \flame models the human head as $\mathcal{F}(\beta, \theta, \psi)$:
\begin{align} 
\begin{split}
    \mathcal{F}(\beta, \theta, \psi) &= \mathbf{LBS}(\mathbf{M}(\beta, \theta, \psi), \mathrm{J}(\beta), \theta, \mathcal{W}) \\
    \mathbf{M}(\beta, \theta, \psi) &= \mathbf{T} + \mathrm{B_s}(\beta) + \mathrm{B_e}(\psi) + \mathrm{B_p}(\theta),  
\label{equ:flame}
\end{split}
\end{align}
where $\mathbf{T}$ is a rest-pose, zero-shape template, $\mathrm{B_s}, \mathrm{B_e}$ and $\mathrm{B_p}$ are shape, expression and pose blendshapes, respectively. $\mathbf{M}$ is the template with blendshape offsets in canonical space. $\mathbf{LBS}$ is the linear blend skinning (LBS) function \cite{loper2023smpl}, that wraps $\mathbf{M}$ to the target pose with skinning weights $\mathcal{W}$ and joints $\mathrm{J}$. The joint locations are defined as: 
\begin{align}
\mathrm{J}(\beta) &= \mathcal{J}(\mathbf{T} + \mathrm{B_s}(\beta)),
\label{equ:joint-mapping}
\end{align}
where $\mathcal{J}$ is a sparse matrix defining how to compute joint locations from mesh vertices. 

For clarity, we define $\kappa = (\beta, \theta, \psi)$, representing shape, pose, and expression, respectively. Let $\mathcal{B} = (\mathrm{B_s}, \mathrm{B_e}, \mathrm{B_p})$ denote the set of blendshapes corresponding to shape, expression, and pose deformations. A rigged parametric model is denoted as $\Omega = (\mathbf{T}, \mathbf{F}, \mathcal{W}, \mathcal{J}, \mathcal{B})$, where $\mathbf{T}$ and $\mathbf{F}$ represent the vertex positions and triangle connectivity, $\mathcal{W}$ the skinning weights, $\mathcal{J}$ the joint definitions, and $\mathcal{B}$ the blendshape basis. This model can be animated or deformed via the control parameters $\kappa$, yielding a posed avatar $\mathcal{F}(\Omega, \kappa)$. We denote by $\bar{\Omega}$ the FLAME model fitted from the generated multi-view observations, and by $\bar{\kappa}$ the identity-neutral, rest-pose configuration (i.e., zero shape, neutral expression, and canonical pose).

Previous works~\cite{Khakhulin2022ROME, yao2021deca, danvevcek2022emoca} typically model diverse 3D head shapes by varying $\kappa$ and adding learned vertex offsets to $\mathcal{F}(\Omega, \kappa)$, while keeping the rigging and expression bases in $\Omega$ fixed across identities. However, due to the limited modeling capacity of $\kappa$ and the fixed topology of the underlying template, these methods often produce overly smooth geometry and struggle to represent complex hairstyles or fine-grained personal details. In contrast, \modelname addresses these limitations by optimizing a personalized $\Omega$ for each input identity, enabling more expressive and detailed reconstructions.

The overview of \modelname is illustrated visually in \cref{fig:pipeline}.  First, six orthogonal RGB images and normal images are generated from the input image using the multi-view diffusion models (Sec.~\ref{sec:multi-view-diffusion}). Next, we deform an initialized \flame mesh $\mathcal{F}(\bar{\Omega},\kappa_{\mathrm{I}})$ to $\mathcal{F}(\Omega^*,\kappa_{\mathrm{I}})$ that accurately aligns with the multi-view normals (Sec.~\ref{sec:geometry}). Finally, we fit the eyeballs and teeth, and generate the UV texture using the multi-view RGB images (Sec.~\ref{sec:texturing}).

\section{Style-Omniscient Sparse-View Generation}
\label{sec:multi-view-diffusion}

As the shape and texture of the head among different styles vary in a wide range, directly regressing the 3D head from a single view is very challenging.
Inspired by the success of 3D generative methods~\cite{tang2025lgm,wu2024unique3d,long2024wonder3d}, we take sparse multi-view images with both appearance and geometry information as the bridge between the single-view portrait and the 3D head.
The generated multiple views typically have high-resolution textures and share quasi-consistent geometries, which are very helpful to achieve high-quality 3D head reconstruction. However, using existing multi-view diffusion models for 3D head reconstruction is suboptimal. Current diffusion priors are trained on general objects rather than being specifically tailored to the head domain, often resulting in less effective head reconstruction. Additionally, there is a lack of large-scale 3D head datasets that cover a diverse range of styles, hairstyles, and accessories.

\subsection{3D Head Dataset}
To build a style-omniscient multi-view generator, the ideal way is to first collect large-scale stylized and as diverse as possible textured 3D heads. 
This is apparently difficult.
\rebuttal{Among the styles covered by publicly available datasets, we observe that anime stands out as the non-realistic style that differs most from real humans.
Anime characters typically feature tiny, sharp noses, large, square eyes, flat faces, simplified hair textures, and a variety of hair accessories (see details in \cref{fig:dataset}).
This observation inspires us to leverage the two highly distinct styles—realistic and anime—to train the generative model, enabling it to imagine and generalize intermediate styles such as oil painting and Chinese ink-and-wash drawing.}
We put our efforts into obtaining more data and finally collect 24$k$ 3D avatars across two styles, anime and realistic, featuring a wide variety of head shapes, hairstyles, expressions, and identities. 
Illustration of our motivation is shown in \cref{fig:dataset}.

For the realistic style, we first collected 9.1$k$ realistic heads, which are 2$k$ from THuman2.0 dataset \cite{tao2021thuman}, 1.8$k$ from 2K2K dataset \cite{han20232k2k} and 5.3$k$ from NPHM dataset \cite{giebenhain2023nphm}.
However, we find that the people in these datasets are predominantly young Asians, and the diversity of hairstyles is limited.
Thus, we further synthesize 2.4$k$ 3D heads with diverse hairstyles, like braids, buns, twists, from UniHair~\cite{zheng2024unihair} and various facial features, like black/white skin, beard, elder age, and wrinkles from the texture maps in FFHQ-UV~\cite{bai2023ffhq}, as a supplement. 
For the anime (non-realistic) style, we directly gathered 13$k$ 3D character models from the Vroid 3D dataset \cite{chen2023panic}. 

For each textured 3D head model, we render 11 groups of images using varying random camera distances and y-axis rotations, with each group containing 6 orthogonal images. 
The camera elevation is fixed at 0, and the azimuths angles are set to $\{\beta, \beta+90^{\circ}, \beta+180^{\circ}, \beta+270^{\circ}, \beta+45^{\circ}, \beta+315^{\circ}\}$, where $\beta$ is randomly sampled from $(-45^{\circ}, 45^{\circ})$. 

\input{figs/dataset}

\subsection{Multi-view Diffusion Model}
Our multi-view image and normal diffusion models share the same network architecture as those in Unique3D \cite{wu2024unique3d}. 
We fine-tune them on the collected 3D head dataset. 
The multi-view image diffusion model $\mathcal{D}_r$ takes a single image $\mathbf{I}\in \mathbb{R}^{256\times256\times3}$ as input and outputs six orthogonal RGB images $\mathcal{I}\in \mathbb{R}^{6\times256\times256\times3}$. 
The normal diffusion model $\mathcal{D}_n$ then takes these images $\mathcal{I}$ as input to generate the corresponding normal maps $\mathcal{N}\in \mathbb{R}^{6\times256\times256\times3}$. 
To enhance visual quality, we employ a single-view super-resolution model to upscale the multi-view images and normal maps by a factor of four, achieving a resolution of $2048\times2048$ while preserving multi-view consistency. 

\section{Animatable 3D Head Reconstruction}
\label{sec:reconstruction}
After generating the multi-view images and normal maps, we use them to reconstruct animatable 3D head avatars.
We first estimate a \flame mesh $\mathcal{F}(\bar{\Omega}, \kappa_I)$ and camera $\pi$ as the initialization, following~\cite{danvevcek2022emoca}.
Then we carefully design the optimization process to deform $\mathcal{F}(\bar{\Omega}, \kappa_I)$ to the personalized $\mathcal{F}(\Omega^*, \kappa_I)$ and texture the head mesh.

Specifically, high-quality textured head results should have a shape that fits the normal maps $\mathcal{N}$ as accurate as possible, while preserving parametrization and rigging, such that the avatar can be easily animated via $\kappa$ as the original \flame.
Achieving this is non-trivial.
We observe that fairness and accuracy cannot be achieved simultaneously in the shape optimization of \flame.
Even with varying $\kappa$ and per-vertex displacement~\cite{Khakhulin2022ROME,yao2021deca,danvevcek2022emoca}, the optimized shape tends to collapse or become over-smoothed, as shown in \cref{fig:topology}. 
To address this, we adopt an iterative approach as illustrated in \cref{fig:pipeline} (b), where the optimization of personalized $\Omega_{\mathbf{I}}$ involves the following steps: (1) semantic template deformation $\mathbf{T} \to$ {$\mathbf{T'}$}, where $\mathbf{T}$ and {$\mathbf{T'}$} have the same number of vertices; (2) remeshing and rig interpolation $\Omega \to $ {$\Omega'$}; and (3) iteratively looping steps (1) and (2).

During the deformation process, parametrization and rigging are preserved through constraints related to facial landmarks and head parsing.
After obtaining the 3D head shape, we generate its corresponding head texture from $\mathcal{I}$, and optimize eyeballs and teeth.

\input{figs/topology}
\subsection{Template Deformation}
\label{sec:geometry}

Given multi-view normal images $\mathcal{N}$, the initial mesh $\mathcal{F}(\bar{\Omega}, \kappa_I)$, camera $\pi$, landmarks $\mathbf{L} \in \mathbb{R}^{68\times 2}$ detected from the input $I$ using \cite{bulat2017far} and head parsing maps $\mathcal{P} \in \mathbb{R}^{3\times h \times w\times3}$ obtained via \cite{dinu2022faceparsing}, we iteratively update the template vertices $\mathbf{T}$ using three losses: reconstruction loss $\mathcal{L}_{\mathrm{rec}}$, 
semantic loss $\mathcal{L}_{\mathrm{sema}}$, and landmark loss $\mathcal{L}_{\mathrm{lmk}}$: 
\begin{align}
\begin{split}
\mathcal{L} &= \lambda_{\mathrm{rec}} \mathcal{L}_{\mathrm{rec}} +  \lambda_{\mathrm{sema}} \mathcal{L}_{\mathrm{sema}} +  \lambda_{\mathrm{lmk}} \mathcal{L}_{\mathrm{lmk}}, 
\end{split}
\end{align}
where $\lambda_*$ represents the weights of the respective losses. 
$\mathcal{L}_{\mathrm{rec}}$ aims to align $\mathcal{F}(\Omega, \kappa_{\mathrm{I}})$ with $\mathcal{N}$. 
$\mathcal{L}_{\mathrm{sema}}$ guides the deformation of the hair, face, and neck, while
$\mathcal{L}_{\mathrm{lmk}}$ focuses on preserving the structure of the eyes, nose, lips, and jaw, and keeping the template as symmetric as possible. 

\qheading{Reconstruction Loss}
We use the normal image as the target to deform the template mesh and apply Laplacian smoothing to regularize the surface. The normal loss computes the difference between the target normal maps $\mathcal{N}$ and the rendered normal maps $\mathbf{n}$: 
\begin{align}
\begin{split}
\mathcal{L}_{\mathrm{rec}} &=\mathcal{L}_{\mathrm{norm}} + \lambda_{\mathrm{lap}} \mathcal{L}_{\mathrm{lap}}, \\
\mathcal{L}_{\mathrm{norm}}(\mathcal{N}, \mathbf{n}) &= \sum_{\mathbf{k} \in \upsilon_{\mathbf{n}} } \lambda_{\mathrm{MSE}}^{\mathbf{k}} \norm{\mathcal{N}_{\mathbf{k}}-\mathbf{n_k}}_{2}^{2},
\end{split}
\end{align}
where $\mathbf{n_k}$ is the rendered normal image of the 3D shape $\mathcal{F}(\Omega,\kappa)$ in view $\mathbf{k}$. 
$\lambda^\mathbf{k}_{\mathrm{MSE}}$ is the weight of view $\mathbf{k}$, and
$\upsilon_{\mathbf{n}}$ represents the six views. 

\input{figs/template_optimization}

\qheading{Semantic Loss}
To encourage the deformation which occurs between the semantically corresponding head parts (e.g., face-to-face and hair-to-hair), we utilize the predicted parsing maps to maintain the overall parametrization as \flame. 
Note that we only use three views $\upsilon_{\mathrm{s}}=\{0^\circ, -45^\circ, 45^\circ \}$ for the semantic loss, because the predictions for side/back views by \cite{bulat2017far} are not reliable.  
The semantic loss consists of two terms, i.e., the parsing loss and the eye mask loss:
\begin{align}
\mathcal{L}_{\mathrm{sema}} = \mathcal{L}_{\mathrm{parse}} + \mathcal{L}_{\mathrm{eye}}, 
\end{align}
where the parsing loss $\mathcal{L}_{\mathrm{parse}}$ compute the difference between the parsing map $\mathcal{P}$ and the rendered parsing map $\mathbf{p}$: 
\begin{align}
\mathcal{L}_{\mathrm{parse}}(\mathcal{P}, \mathbf{p}) = \sum_{k\in \upsilon_{\mathrm{s}}}  \norm{
(\mathcal{P}_{\mathbf{k}}-\mathbf{p_k}) \otimes \mathcal{S}
}_{2}^{2}. 
\end{align} 
where $\mathcal{S}$ is the rendered mask of the 3D head excluding the eyeballs.
We do not directly use the eye mask as supervision since the rendered eyeball mask is consistently larger than the observed eye mask. Instead, we push the vertices not belonging to the eyeballs to lie outside the observed eye area. The eye mask loss $\mathcal{L}_{\mathrm{eye}}$ is defined as: 
\begin{align}
\begin{split}
\mathcal{L}_{\mathrm{eye}}(\mathcal{S} , \mathbf{s} ) = \sum_{k\in \upsilon_{\mathrm{s}}}  \norm{\mathcal{S}_{\mathbf{k}}-\mathbf{s_k}}_{2}^{2},
\end{split}
\end{align}
where ${\mathcal{S}}$ is the rendered mask of the 3D shape $\mathcal{F}$ excluding the eyeballs, while  $\mathbf{s}$ denoting the pseudo ground-truth from the parsing.

\qheading{Landmark Loss} The landmark loss is defined as the sum of the landmark projection loss $\mathcal{L}_{\mathrm{lmkpro}}$ and the canonical landmark symmetry loss $ \mathcal{L}_{\mathrm{lmsym}}$: 
\begin{align}
\begin{split}
\mathcal{L}_{\mathrm{lmk}} &= \mathcal{L}_{\mathrm{lmkpro}} + \mathcal{L}_{\mathrm{lmsym}}. 
\end{split}
\end{align}
$\mathcal{L}_{\mathrm{lmsym}}$ ensures that corresponding pairs of canonical landmarks on opposite sides of the face are symmetric with respect to the YZ-plane. It is defined as: 
\begin{align}
\begin{split}
\mathcal{L}_{\mathrm{lmksym}} &= \frac{1}{N} \sum_{i=1}^{N} \| \hat{\mathbf{L}}_{\mathrm{cano}}^{i} - \mathbf{R}(\hat{\mathbf{L}}_{\mathrm{cano}}^{j}) \|^2, \\
 \hat{\mathbf{L}}_{\mathrm{cano}} &= \mathcal{J}_{\mathrm{lmk}} (\mathbf{T}+\mathrm{B_s}(\beta)), 
 \end{split}
\end{align}
where $ \hat{\mathbf{L}}_{\mathrm{cano}}^{i}$ and $\hat{\mathbf{L}}_{\mathrm{cano}}^{j}$ represent a pair of symmetric landmarks in canonical space ($\kappa=\bar{\kappa}$), $\bar{\kappa}$ denotes zero shape and rest pose. 
$\mathbf{R}$ is the reflection transformation about the \( YZ \)-plane. $\hat{\mathbf{L}}_{\mathrm{cano}}$ is mapped from $\mathbf{T}$ as Eq. (\ref{equ:joint-mapping}) using the landmarks mapping matrix $\mathcal{J}_{\mathrm{lmk}} \in \mathbb{R} ^ {68\times |\mathbf{T}|}$ provided in \cite{FLAME:SiggraphAsia2017}. 

The landmark projection loss $\mathcal{L}_{\mathrm{lmkpro}}$ computes the distance between the projected landmarks $\Delta(\hat{\mathbf{L}},\pi^{\mathrm{front}}) \in \mathbb{R}^{68\times2}$ and  the image landmarks $\mathbf{L}$:
\begin{align}
\begin{split}
\mathcal{L}_{\mathrm{lmkpro}} &= ||\mathbf{L} - \Delta(\hat{\mathbf{L}},\pi^{\mathrm{front}})||_2^2 , \\
\hat{\mathbf{L}} &= \mathbf{LBS} (\hat{\mathbf{L}}_{\mathrm{cano}}, \mathrm{J}, \theta, \mathcal{W})
\end{split}
\end{align}
where $\Delta$ is the projection operation, $\pi^{\mathrm{front}}$ denotes the input view camera, and $\hat{\mathbf{L}} \in \mathbb{R}^{68\times 3}$ are the facial landmarks of $\mathcal{F}(\Omega,\kappa_{\mathrm{I}})$. $\hat{\mathbf{L}}$ is warped from the canonical landmark using the LBS function.    

\subsection{Topology Correction}

The deformation process can result in topological issues, such as large triangles, reversed face normals, and twisted faces.
To avoid negatively impacting further optimization, we introduce a remeshing operation and rig interpolation after each deformation step.
We remesh the template mesh from $\mathbf{T} \in \mathbb{R}^{\mathrm{N}\times3}, \mathbf{F} \in \mathbb{R}^{\mathrm{M}\times 3}$ to $\mathbf{T}' \in \mathbb{R}^{\mathrm{N'}\times3}, \mathbf{F}' \in \mathbb{R}^{\mathrm{M'}\times 3}$. 
The remeshing operation consists of three steps: (1) subdividing large triangles with edges larger than $\epsilon$, (2) flipping inconsistent triangles, and (3) removing incorrect triangles. 

As the topology of the template mesh changes, the skinning weights $\mathcal{W}\in \mathbb{R}^{\mathrm{N}\times |\mathrm{J}|}$, blendshapes  $\mathcal{B}\in \mathbb{R}^{500\times |\mathrm{J}|}$,  and joint mapping matrix $\mathcal{J} \in \mathbb{R}^{|\mathrm{J}|\times \mathrm{N}}$ must also be updated accordingly. 
For each newly added vertex, $\mathcal{W}$ and $\mathcal{B}$ are computed by interpolating the corresponding values from neighboring vertices along the edge. However, $\mathcal{J}$ cannot be interpolated in the same way, as this would change the location of joints defined by Eq. (\ref{equ:joint-mapping}). To preserve the joint positions, the joint regressor matrix $\mathcal{J}$ should be updated as:
\begin{align}
\mathcal{J} = \mathrm{J} (\mathbf{T} + \mathrm{B_s}({\beta}))^{-1},
\end{align}
where $\mathrm{J}$ represents the canonical joints obtained from Eq. (\ref{equ:flame}) using the newly deformed template. This approach assumes that the joints in canonical space remain unchanged after interpolation.   

\subsection{Eyeballs, Teeth and Textures}

\qheading{Eyeball Optimization} To fit the eyeballs, we start with the normalized eyeballs from \flame and optimize for a shared radius $\mathbf{r}$ and the centers $\mathbf{c}$ of both eyeballs. Specifically, we render the eye mask from several viewpoints of the reconstructed head, which guides the optimization of $r$ and $c$. 

\qheading{Teeth Alignment}   
We align and register a 3D teeth template into the mouth region of the reconstructed 3D head. This process is straightforward, as the first 5023 vertices of our reconstructed head mesh follow the same ordering as \flame.  
Consequently, the teeth template can be precisely scaled and positioned within the mouth based on the corresponding vertices. 
The blendshape and joint regressor are set to zero. Skinning weights for the upper teeth are assigned to the head, while those for the lower teeth are assigned to the jaw. 
The same teeth template is shared across different styles. Thus, the styles of the avatar and teeth may not be consistent. Fortunately, since we produce mesh-based models, the teeth can be easily edited or replaced by artists.

\qheading{Texture Generation} 
After reconstruction, we generate the mesh texture of the optimized head and eyeballs using the multi-view images $\mathcal{N}$. The UV map from \flame is adopted as the initial map and interpolated during mesh interpolation. The texture of the teeth template (for teeth and gum) is then merged into the interpolated UV. The textured map can be obtained by blending colors from different views based on surface normals. For invisible areas, the texture is inpainted by dilating the texture map, as the UV map is continuous. This continuity enables the editing of facial texture, as shown in \cref{fig:tex_edit}.

\label{sec:texturing}

%% file: figs/dataset.tex
\begin{figure}
\centering
    \includegraphics[width=0.5\textwidth]{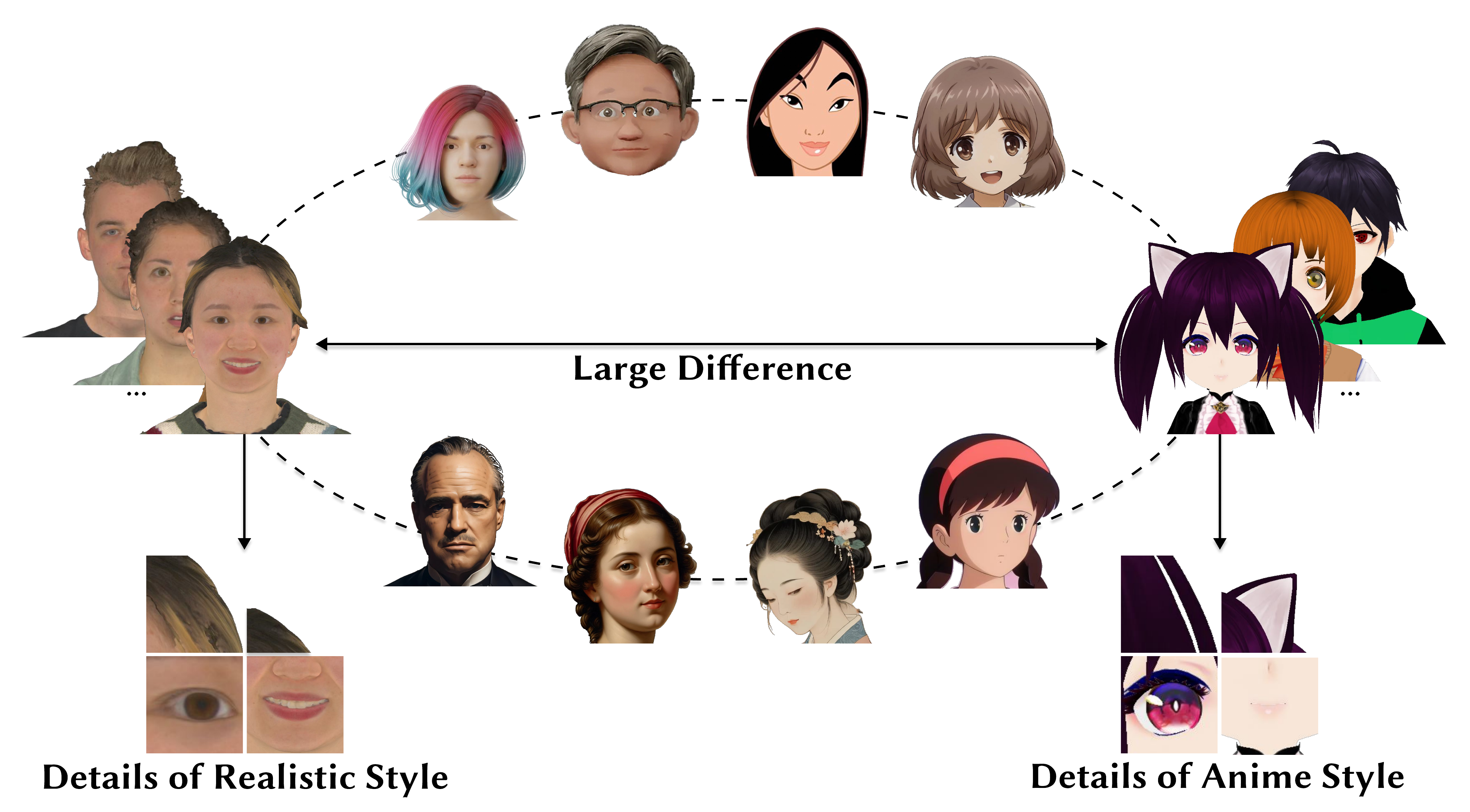}
    \caption{  
        3D Head dataset. The idea is to train the diffusion module with only two extreme styles, i.e., realistic and anime (non-realistic), and generalize to unseen intermediate styles.
    }
    \vspace{-0.3cm}
    \label{fig:dataset}
\end{figure}

%% file: figs/topology.tex
\begin{figure}
\centering
    \includegraphics[width=0.48\textwidth]{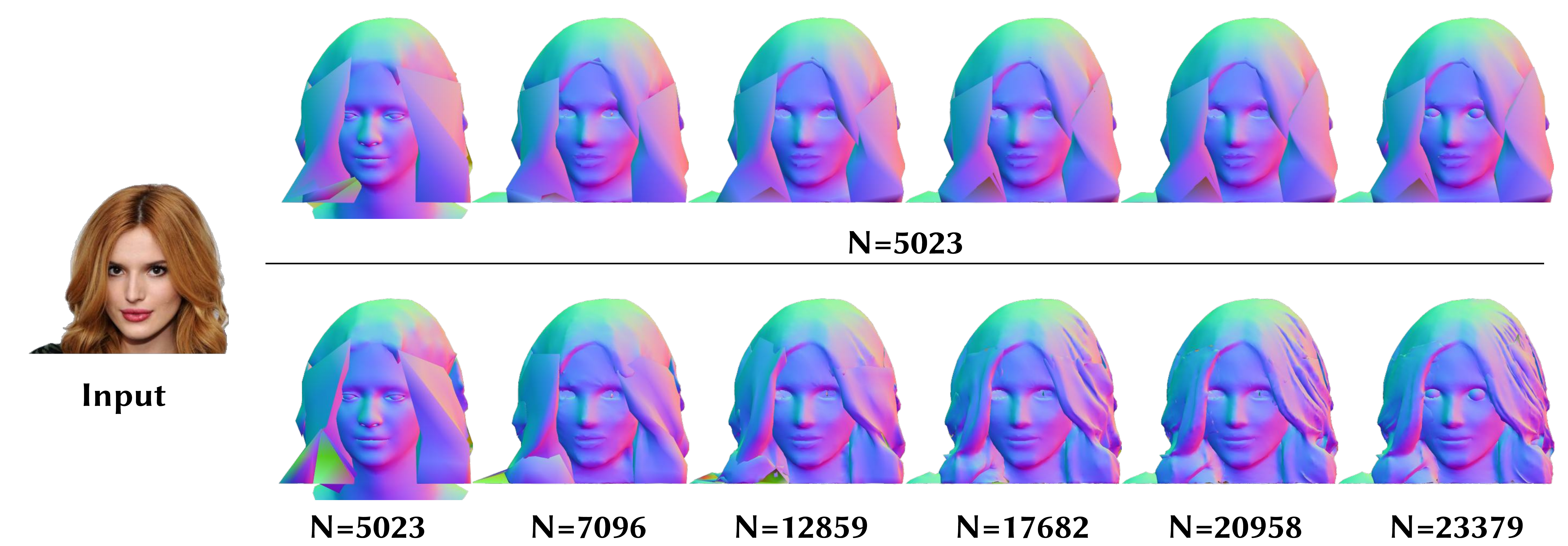}
    \caption{  
        Motivation for topology correction. The top and bottom rows show results with and without topology correction. In this example, the optimized mesh fails to reconstruct the geometric details of the hair and face without topology correction. Due to the significant deformation of hair starting from the FLAME scalp, there is a tendency for undesired twists and collapses, as highlighted in the red boxes.
    }
    \vspace{-0.2cm}
    \label{fig:topology}
\end{figure}

%% file: figs/template_optimization.tex
\begin{figure}%
\centering
    \includegraphics[width=0.4\textwidth]{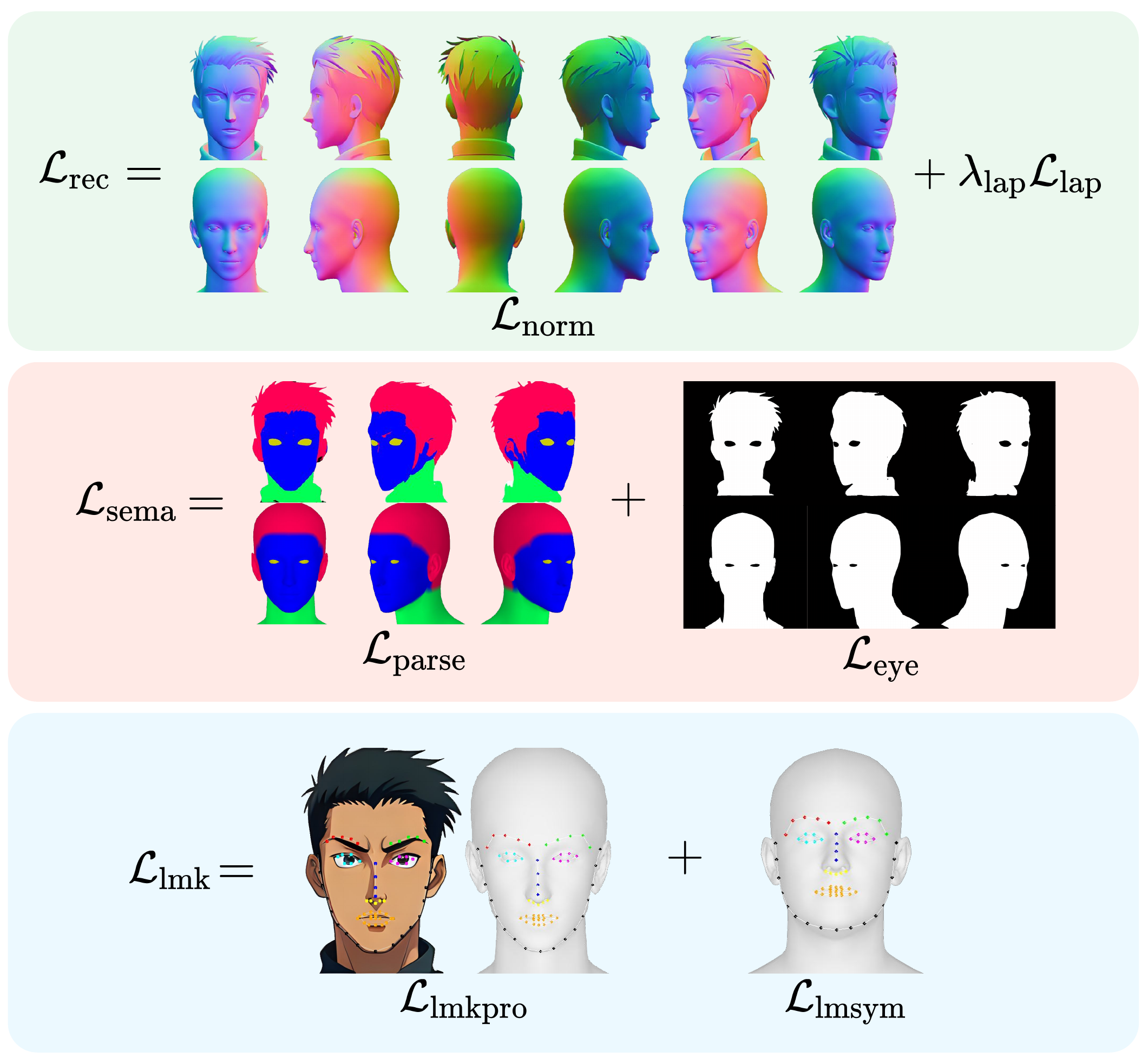}
    \vspace{-1em}
    \caption{
        Template optimization losses. Illustrations of reconstruction, semantic, and landmark losses to template deformation.
        }  
        
        \vspace{-1em}
    \label{fig:template_optimization}
\end{figure}

%% file: sec/5_experiments.tex
\input{figs/metric}

\section{Experiments}
\label{sec:experiments}

\subsection{Implementation Details}
\label{sec:implement-details}


\qheading{Diffusion Training} We adopt the same network architecture for both the multi-view image and normal diffusion models as in \cite{wu2024unique3d}. The entire training process takes approximately 6 days on 7 NVIDIA A6000 GPUs, with a batch size of 48 per GPU under float16 precision. Both the input and output image resolutions are set to 256x256. The AdamW optimizer is used with a learning rate of $4 \times 10^{-4}$, a weight decay of 0.05, and betas of (0.9, 0.999). The total number of iterations is set to 40K. 
We initialize our networks with the pretrained weights from Unique3D~\cite{wu2024unique3d}.

\qheading{Mesh Reconstruction} Each template deformation involves 800 iterations, with the mesh interpolated six times. During the first epoch, the face and neck vertices are fixed, and only the hair vertices are optimized. This helps the mesh hair to align more accurately with the target. For interpolation, $\epsilon$ is set to $5 \times 10^{-4}$, while the shortest edge length is $5 \times 10^{-5}$.

\qheading{Inference Time} It takes approximately 6 minutes to generate an animatable textured 3D avatar from a single image.
Specifically, the pre-processing step takes about 1.5 minutes, including landmark detection, face parsing, FLAME initialization, and the generation of multi-view images and normal maps.
The textured head reconstruction then requires an additional 4.5 minutes, covering head and eyeball optimization, tooth alignment, and texture generation. 

\subsection{Comparisons}
\qheading{Qualitative Comparisons} 
We compare our method with diffusion-based methods and parametric-based methods. Diffusion-based methods include Unique3D \cite{wu2024unique3d}, LGM \cite{tang2025lgm}, and Wonder3D \cite{long2024wonder3d}. As shown in \cref{fig:compare_diffusion}, the 3D heads produced by \modelname exhibit superior visual quality and better consistency, particularly in the back and side views. Furthermore, \modelname produces more reasonable and natural head shapes compared to those methods trained with general objects.
The parametric-based methods include ROME \cite{Khakhulin2022ROME}, as well as commercialized products such as Itseez3D\footnote{\url{https://itseez3d.com/}} and AvatarNeo\footnote{\url{https://avatarneo.com/}} \cite{luo2021normalized,hu2017avatar}. As shown in \cref{fig:comp_parametric}, our method can produce high-quality and image-aligned 3D reconstruction of the full head, offering a more realistic and detailed reconstruction compared to these alternatives.
ROME and Itseez3D fail to generate the faithful back view.
AvatarNeo \cite{luo2021normalized} outputs retrieval-like hairstyles and cannot capture exaggerated expressions, as shown in the first example of \cref{fig:comp_parametric}.

\qheading{Quantitative Comparisons}   
To evaluate the performance of our method, we collected 40 head scans, consisting of 20 real scans from \cite{giebenhain2023nphm} and 20 synthetic scans in various styles like Anime and CG. We assess the results using five metrics: Peak Signal-to-Noise Ratio (PSNR), Structural Similarity Index Measure (SSIM), Learned Perceptual Image Patch Similarity \cite{zhang2018perceptual} (LPIPS), Cosine Similarity (CSIM), and Fréchet Inception Distance \cite{10.5555/3295222.3295408} (FID). For each mesh, we render 8 views across azimuth angles at 45-degree intervals to compute the metrics with the ground-truth images. For the CSIM metric, we utilize the face recognition model \cite{deng2019arcface} and frontal views as the reference for comparison. 
\rebuttal{Please note that we use 60 views per mesh when computing FID to ensure stability. We also compute 3D metrics, i.e., chamfer distance (CD), IoU, and $F_1$ score, against the ground-truth meshes as a supplement. }
The results, presented in Tab. \ref{table:metrics}, show that \modelname consistently outperforms the compared methods across PSNR, SSIM, LPIPS, FID, CD, IoU, and $F_1$ score, demonstrating its superior performance. 
\rebuttal{Other methods do not perform well, because: 1) GAN-based methods such as PanoHead~\cite{An2023PanoHead} and SphereHead~\cite{li2024spherehead} are trained only on realistic data;
2) Diffusion-based approaches, i.e., Wonder3D~\cite{long2024wonder3d}, LGM~\cite{tang2025lgm}, and Unique3D~\cite{wu2024unique3d}, are trained for general objects generation and have not been tailored to the head domain;
3) ROME~\cite{Khakhulin2022ROME} is limited to generating only the frontal head and fails to reconstruct the back (check the comparisons from different views in the supplementary materials).}

\input{figs/ablation}

\subsection{Ablation Study} 
We ablate the mesh reconstruction
of \modelname on three key components: semantic loss $\mathcal{L}_{\mathrm{sema}}$, landmark loss $\mathcal{L}_{\mathrm{lmk}}$, and remeshing operation. 
\cref{fig:ablation} presents the ablation comparisons, leading to several important observations. 
Landmark loss is key to preserving facial animation features such as the lips, jawline, and eyebrows. Semantic loss maintains the overall head structure, especially for long or drooping hairstyles—without it, face-boundary vertices may shift into the hair region. Finally, remeshing is critical for accurately reconstructing complex surfaces like long hair and headwear; without it, the surface becomes overly smooth due to limited vertex density and fixed topology.

\subsection{More Results and Applications} 

\qheading{Animation} 
Our generated 3D head avatars are fully animatable using \flame parameters. 
\cref{fig:animation} showcases 3D avatars in various styles animated with diverse expressions, demonstrating the capability of \modelname to produce expressive and stylized animated avatars.
\qheading{Texture Editing} As our UV map is interpolated from \flame UV, its continuity enables texture editing of the reconstructed 3D heads, as shown in \cref{fig:tex_edit}.

\qheading{Diversity}    
As illustrated in \cref{fig:diverse_avatar}, our method demonstrates a high level of diversity in generating a wide range of avatar styles, including realistic, Pixar, Disney, and anime-inspired designs. In addition to style variation, our geometry optimization approach is highly adaptable, effectively capturing different hairstyles, headwear, and other intricate details. 

\input{figs/animation}



\subsection{Limitation and Discussion}  
\modelname relies on several dependencies, including FLAME estimation, landmark detection, and head parsing. Although these methods generally perform well for realistic avatars, their performance is significantly less effective for certain stylized inputs, such as anime, cartoon, or heavily exaggerated artistic styles. This limitation can result in inaccurate landmark detection, suboptimal head-parsing segmentation, and misaligned FLAME estimates, which in turn affect the quality of the final 3D reconstruction. As a result, the avatars generated may show a distorted head structure, as illustrated in \cref{fig:failure}. Addressing these issues would require either improving the robustness of the dependencies to handle a wider variety of styles or developing more style-agnostic alternatives tailored specifically for stylized avatar reconstruction. 

\rebuttal{The limited resolution of the output of diffusion models restricts the quality of the final results.
During topology correction, our choice of the number of vertices is chosen to match the effective resolution of the normal prediction of the input image. For certain cases, such as sharp edges of hair, adding more vertices would indeed ensure less blurry normal rendering, but would also introduce extra computation.
To improve reconstruction quality, one approach is to increase the resolution of the output images and normal maps from multiview diffusion models (currently $256\times256$), which requires more advanced hardware (GPU). Higher-quality data can bring about further improvement.}

%% file: figs/metric.tex
\begin{table}[h]
\centering
\scriptsize
\setlength{\tabcolsep}{3.5pt}
\renewcommand{\arraystretch}{1.2}
\caption{
\rebuttal{Quantitative comparisons with existing single-view 3D head avatar reconstruction methods across different styles. }
} 
\vspace{-1.0 em}
\resizebox{0.97\linewidth}{!}{
\centering
\begin{tabular}{l|ccccc|ccc}
Method                                            & PSNR $\uparrow$  & SSIM $\uparrow$ & LPIPS $\downarrow$ & CSIM $\uparrow$ & FID $\downarrow$ & CD $\downarrow$ & IoU $\uparrow$ & $F_1$ $\uparrow$  \\ \hline
ROME    & 12.60       & 0.7257     & 0.3290      & 0.6202 & 131.4 & - & - & - \\
PanoHead    & 3.372      & 0.2998      & 0.6044       & 0.2507 & 116.3 & - & - & - \\
SphereHead    & 6.683      & 0.5404      & 0.4703       & 0.4935 & 119.4 & - & - & - \\
Wonder3D  & 16.17 & 0.7806 & 0.2298  & 0.6484 & 102.2 & 5.729 & 0.2395 & 0.3367 \\
LGM            & 16.01 & 0.7914 & 0.2581  & \underline{0.8511} & 104.0 & - & - & - \\
Unique3D    & \underline{17.09} & \underline{0.7947} & \underline{0.2064}  & \textbf{0.8995} & \underline{70.64}  & \underline{4.991} & \underline{0.2409} & \underline{0.3777}\\ \hline
\textbf{Ours}   & \textbf{17.53}       & \textbf{0.7958}      & \textbf{0.1968}       & 0.8268      & \textbf{58.44}  & \textbf{2.880} & \textbf{0.3035} & \textbf{0.5108}   
\end{tabular}}
\vspace{-3.0 em}
\label{table:metrics}
\end{table}


%% file: figs/ablation.tex
\begin{figure}
    \centering
    \includegraphics[width=0.48\textwidth]{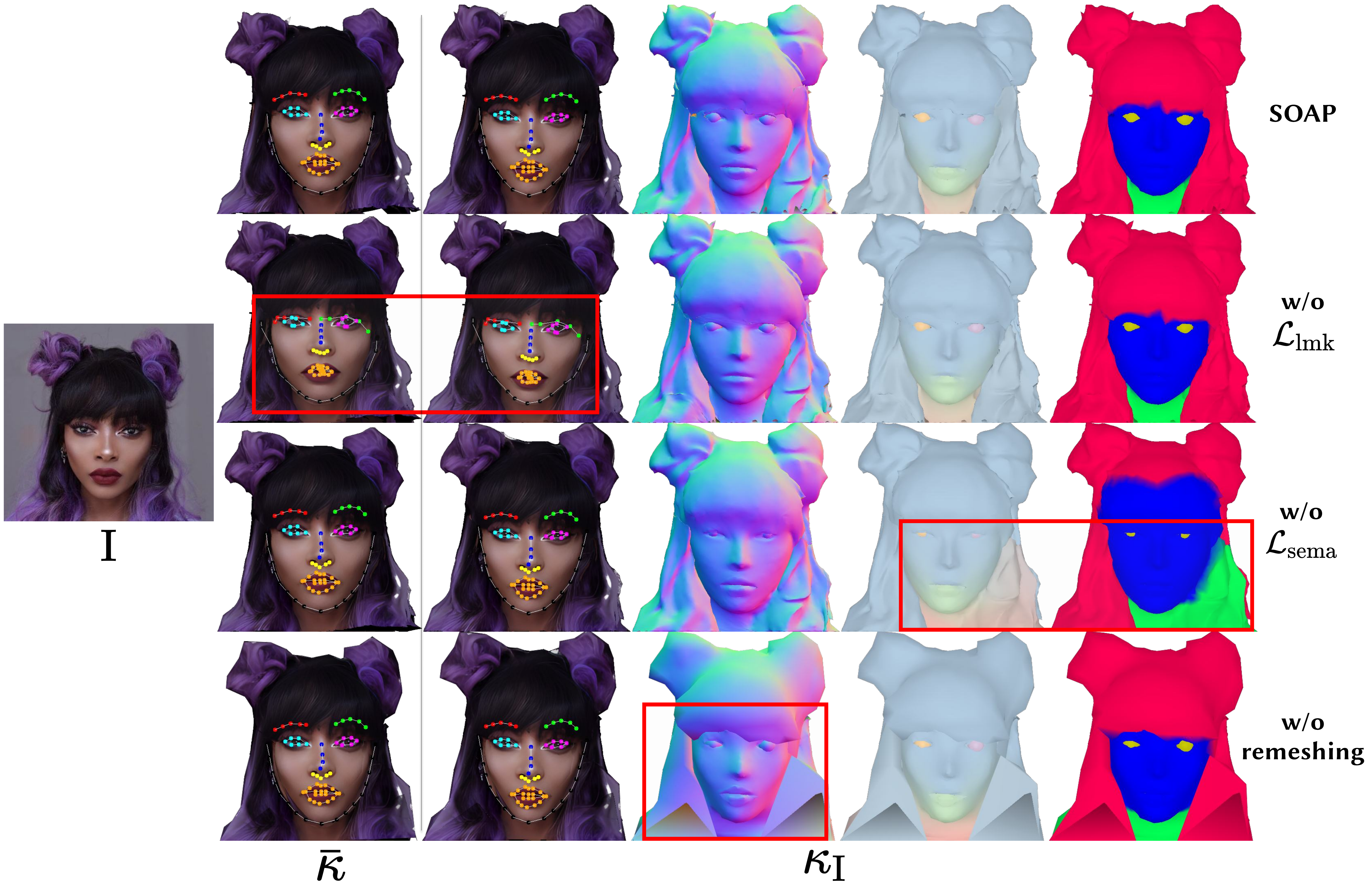}
    \vspace{-0.47cm}
    \caption{  
        Ablation study. Impact of landmark loss, semantic loss, and remeshing on mesh reconstruction.   
    }
    \vspace{-0.47cm}
    \label{fig:ablation}
\end{figure}

%% file: figs/animation.tex
\begin{figure}%
\centering
    \includegraphics[width=0.45\textwidth]{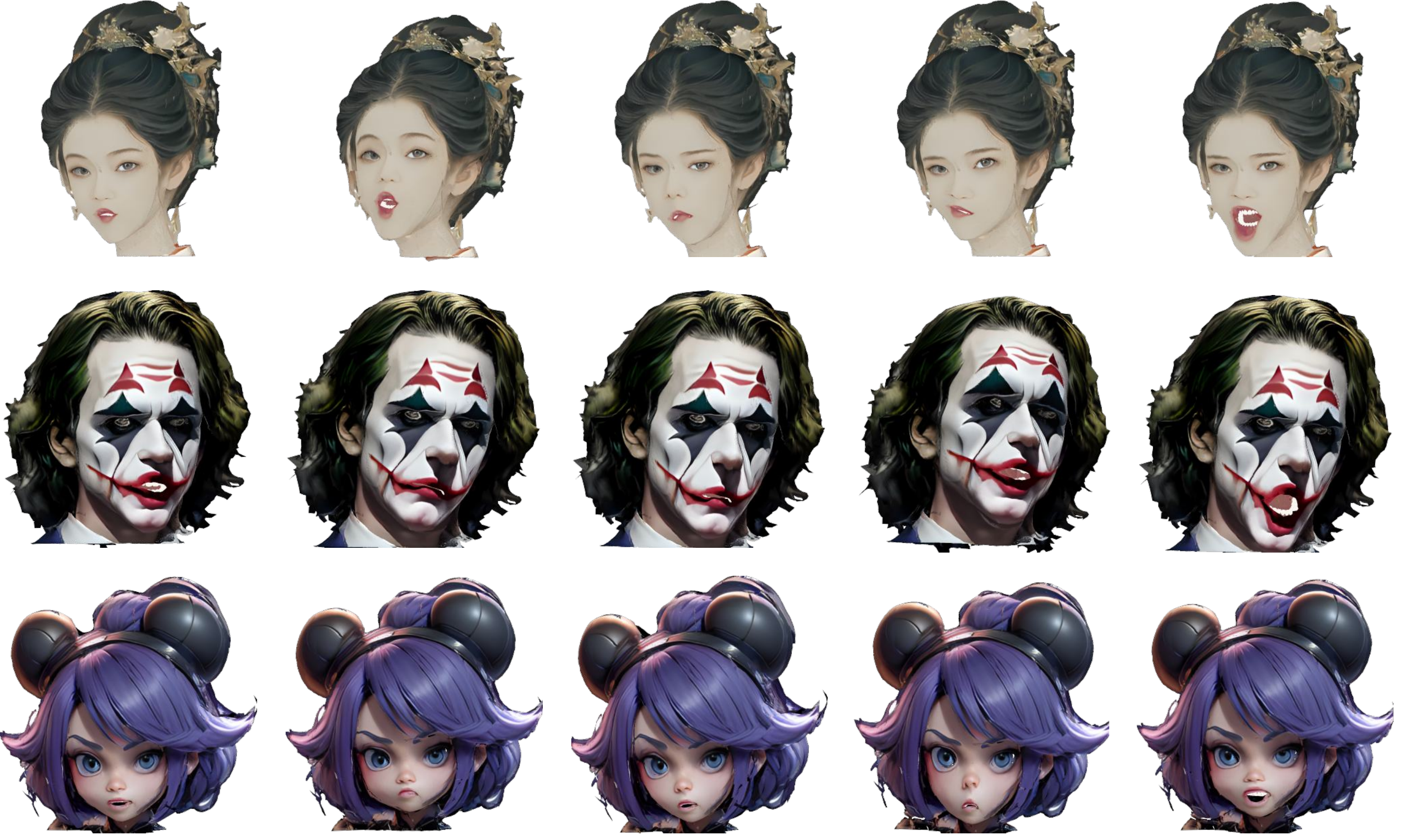}
    \vspace{-0.3cm}
    \caption{
        Animation. Expressive and stylized 3D avatars animated using FLAME parameters. 
    }
    \vspace{-0.5cm}
    \label{fig:animation}
\end{figure}

%% file: sec/6_conclusion.tex
\section{Conclusion}

In this work, we present a novel approach, \modelname, that enables the modeling of style-agnostic, animatable 3D head avatars from single-view portraits. We demonstrate that multi-view diffusion models trained on a limited dataset with two extreme styles can generalize to a wide range of intermediate head styles.
To reconstruct animatable 3D heads from sparse views, we have designed an optimization method that includes parametric template head deformation and topology correction in each iteration. With the integration of topology correction and semantic constraints, our optimization process can effectively manage the significant variability in styles, resulting in high-quality outputs for both geometry and texture.
Extensive experiments have confirmed the effectiveness of our approach for single-view 3D head reconstruction across various styles. The reconstructed 3D heads can be easily animated using head pose and expression parameters, or directly through video input, and can be further customized by adjusting the shape parameters.

%% file: sec/acknowledgement.tex
\section{Acknowledgment}
This work was supported by the Metaverse Center Grant from the MBZUAI Research Office. Yuliang Xiu was supported by the Research Center for Industries of the Future (RCIF) at Westlake University and the Westlake Education Foundation.

%% file: figs/combine_2_figures.tex
\begin{figure*}[t]
    \centering
    \begin{minipage}[t]{0.48\linewidth}
        \centering
        \includegraphics[width=0.95\textwidth]{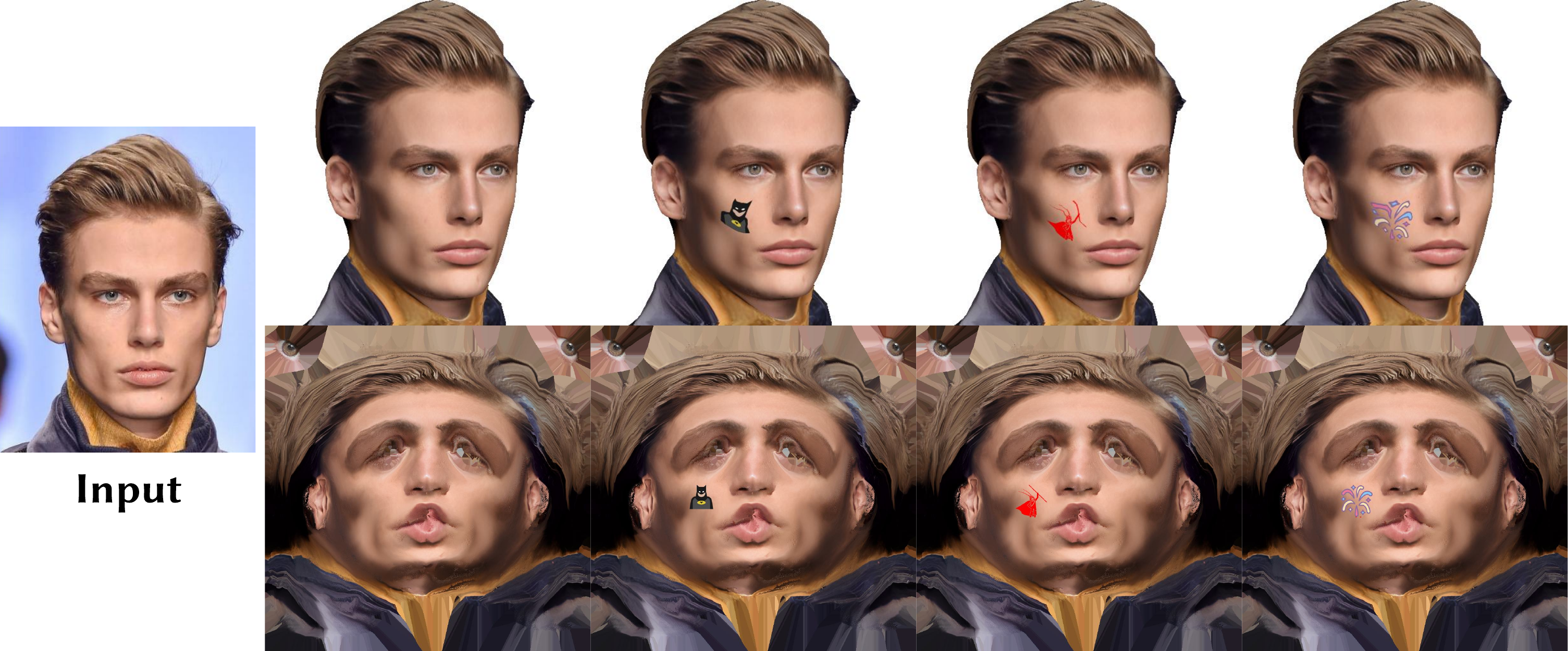}
        \caption{  
            Editable textures are enabled by the interpolated UV map. 
        }
        \label{fig:tex_edit}
    \end{minipage}
    \hfill
    \begin{minipage}[t]{0.48\linewidth}
        \centering
        \includegraphics[width=0.95\textwidth]{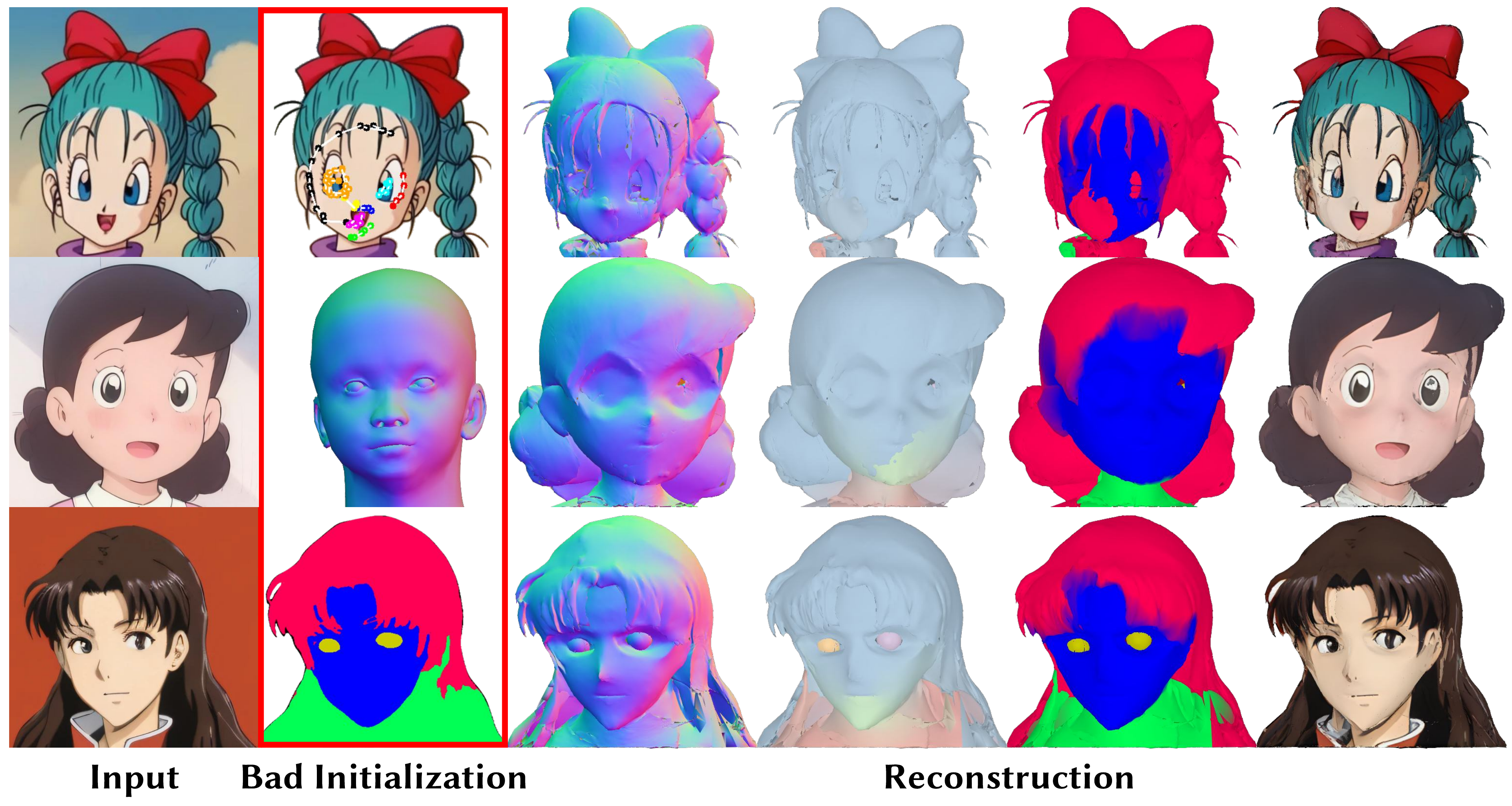}
        \caption{  
            Failure cases. The reconstruction results with incorrect landmarks, erroneous \flame initialization, and inaccurate head parsing.    
        }
        \label{fig:failure}
    \end{minipage}
\end{figure*}

%% file: figs/diverse_avatars.tex
\begin{figure*}[!h]
\centering
    \includegraphics[width=0.95\textwidth]{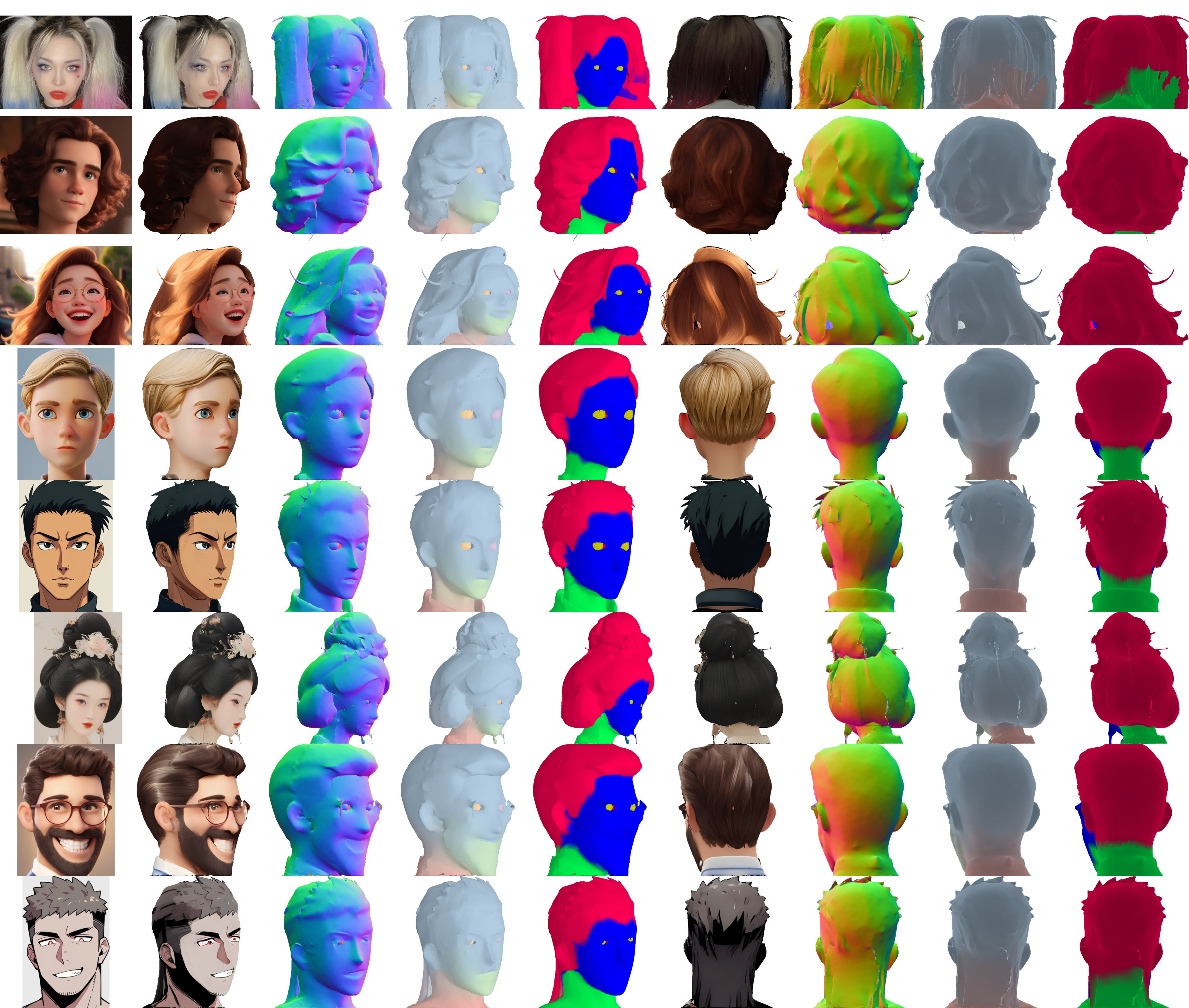}
    \vspace{-0.3cm}
    \caption{
        Diverse avatars. From left to right are the input image, the rendered RGB, normal, skinning weights, and parsing labels.  
    }
    \vspace{-0.5cm}
    \label{fig:diverse_avatar}
\end{figure*}

%% file: figs/comparison.tex
\begin{figure*}%
\centering
    \includegraphics[width=0.92\textwidth]{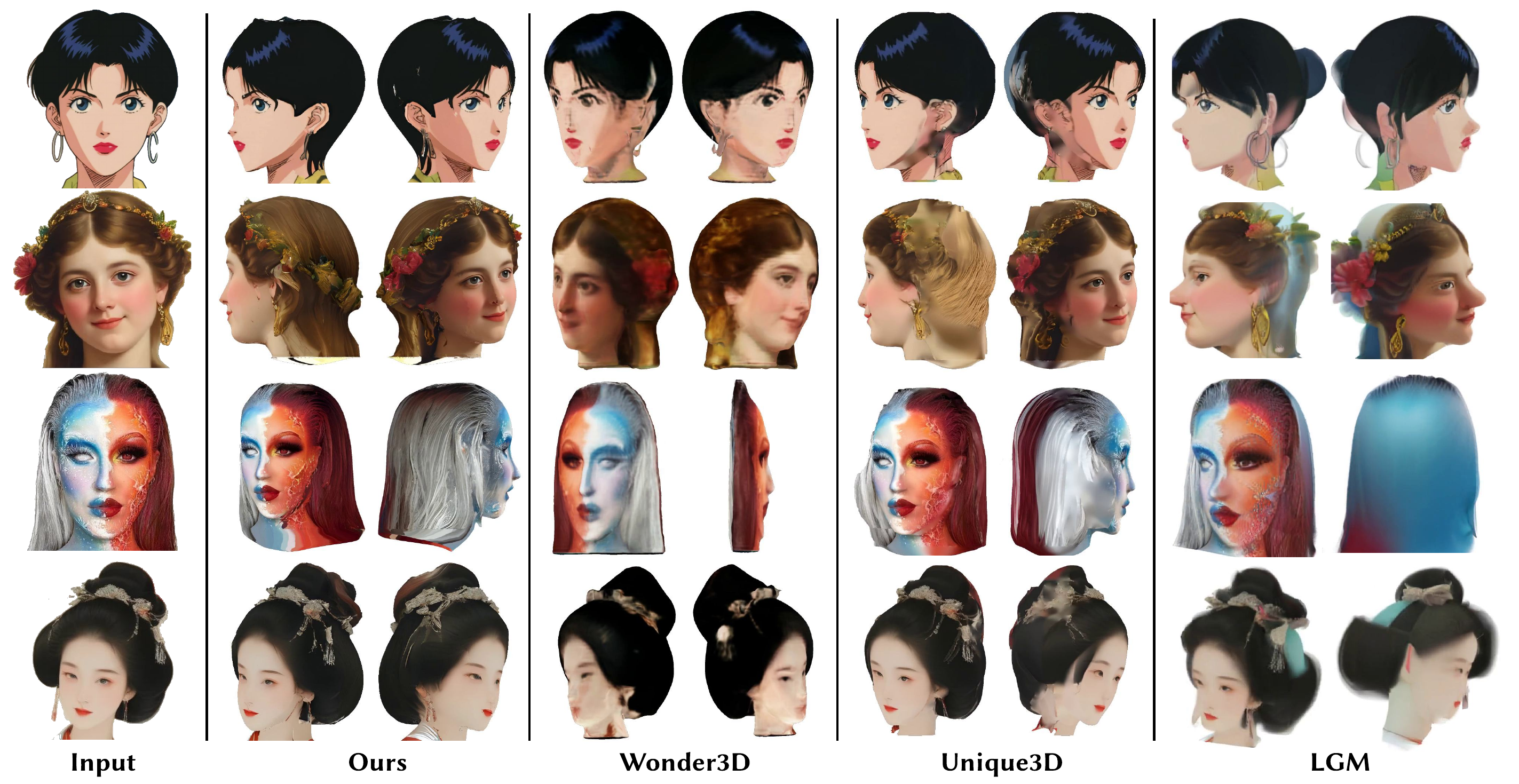}
    \vspace{-0.2cm}
    \caption{
        Qualitative comparisons with diffusion-based methods.  
    }
    \label{fig:compare_diffusion}
\end{figure*}
\begin{figure*}%
\centering
    \includegraphics[width=0.92\textwidth]{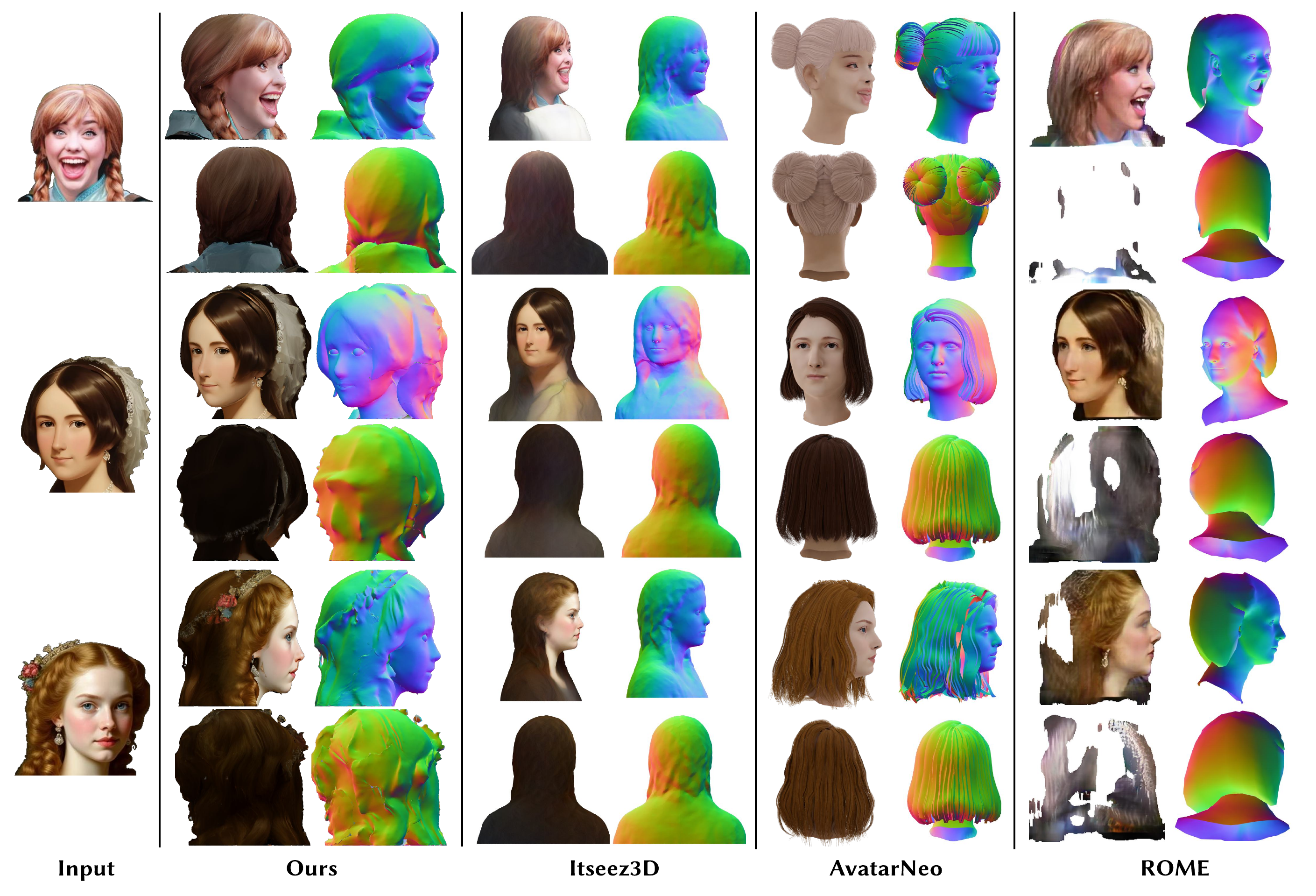}
    \vspace{-0.2cm}
    \caption{
     Qualitative comparisons with parametric-based methods.  
    }
    \label{fig:comp_parametric}
\end{figure*}

%% file: sec/appendix.tex
\begin{appendix}
\label{appendices}

\input{figs/data_gen}
\input{supp/data_example}
\input{figs/comparison_diffusion}

\section{Dataset}

To supplement the diversity of hairstyle and facial features, we generate 2.4$k$ 3D heads with the hairstyles in UniHair~\cite{zheng2024unihair} and facial textures in the FFHQ-UV dataset~\cite{bai2023ffhq}.
As illustrated in \cref{fig:data_gen}, we first select head UV maps in the FFHQ-UV dataset~\cite{bai2023ffhq} with representative facial appearances, such as black/white skin, beard, elder age, and wrinkles.
Then we map selected textures to 3D head models and adapt eyeballs as well as various hairstyles from UniHair~\cite{zheng2024unihair}, like braids, buns, twists, and more.
Importantly, as some hairstyles are partially bald and the texture maps of all heads include black short hair textures across the entire scalp region, we use SAM~\cite{kirillov2023segment} to segment the scalp and replace those pixels with the corresponding skin color for each selected texture map.

\input{figs/comparison_unihair}
\input{figs/comparison_xportrait}
\input{figs/comparison_panic3d}
\input{figs/comparison_spherehead}

\input{figs/rome_tab}

\input{figs/user_study_3d}

\input{figs/user_study_mv}

\section{More Comparisons}
\rebuttal{We provide the discussions and comparisons with Unihair~\cite{zheng2024unihair}, X-Portrait~\cite{xie2024x}, Panic3D~\cite{chen2023panic}, ROME~\cite{Khakhulin2022ROME}, PanoHead~\cite{An2023PanoHead} and SphereHead~\cite{li2024spherehead} in this section.}

\rebuttal{\qheading{UniHair} Unihair focuses solely on hair reconstruction instead of the full head reconstruction. It is tailored to the realistic portraits and often fails on stylized ones, as shown in~\cref{fig:compare_unihair}.}

\rebuttal{\qheading{X-Portrait} 
X-Portrait is a 2D reenactment technique, while SOAP generates fully 3D animatable avatars.
Visual comparisons are shown in~\cref{fig:compare_xportrait}.
As observed in the second row of~\cref{fig:compare_xportrait}, X-Portrait sometimes produces more natural expressions and more accurate eyeball movements in certain frames. 
This is primarily because SOAP is constrained by the FLAME blendshapes and the limitations of eye-tracking performance.
On the other hand, SOAP enables multi-view rendering (as shown in the last four rows of~\cref{fig:compare_xportrait}) and supports traditional editing workflows through the use of 3D assets and rendering.
In contrast, X-Portrait remains a purely 2D approach.
Depending on the application requirements, one method may be more suitable than the other.
}



\rebuttal{ \qheading{Panic3D} We summarize the differences between SOAP and Panic3D along three aspects: (1) Animation: Panic3D is a novel view synthesis method that does not support animation; (2) Style: Panic3D is limited to the anime style, whereas SOAP supports multiple styles; (3) Quality: Panic3D relies on triplane and NeRF-based reconstruction, leading to low-resolution and blurry outputs. Visual comparisons are provided in~\cref{fig:compare_panic3d}. We note that our comparisons are limited to examples shown in their paper, as the code for image-conditioned inference in Panic3D has not been released. }

\rebuttal{ \qheading{ROME} We provide a quantitative comparison across different views for ROME~\cite{Khakhulin2022ROME} in~\cref{tab:rome}. Since ROME is primarily effective at reconstructing the frontal head, its performance is significantly better on frontal views (azimuths of $0^\circ$, $45^\circ$, and $315^\circ$) compared to side and back views. In contrast, our method consistently achieves superior results across all views. }

\rebuttal{\qheading{PanoHead and SphereHead} Visual comparisons with PanoHead~\cite{An2023PanoHead} and SphereHead~\cite{li2024spherehead} are presented in~\cref{fig:compare_gan}. As these GAN-based methods are trained primarily on realistic data, they struggle to perform well on stylized portraits.}

\section{User Study}
\rebuttal{For the user study on the final results, we render 360-degree videos of the reconstructed 3D models and present each volunteer with five examples, following the protocol of Unique3D~\cite{wu2024unique3d}. Each example includes the input images and video samples from all methods, covering the styles of real human, joker, anime, oil painting, and 3D cartoon. Participants rate the videos based on three criteria—view consistency, ID consistency, and overall quality—using a 1–5 scale (with higher scores indicating better performance). The average scores from 30 volunteers are reported in Tab.~\ref{table:user_study_3d}, where our method consistently receives higher ratings across all criteria. } 

\rebuttal{The quantitative evaluation of view and ID consistency for multi-view diffusion is presented in Tab.~\ref{table:user_study_mv}. The evaluation follows a similar protocol to the user study for 3D results, with the key difference being that participants are shown only the generated multi-view images and normal maps from the diffusion modules, rather than videos. Our diffusion modules outperform those used in Unique3D, as they are specifically tailored to the domain of human heads. }

\section{Evaluation of Diffusion Model}
\qheading{Comparisons on Different Models} We compare our 6-view diffusion model and \modelname (4-view), which is trained with four orthogonal perspectives, against Unique3D~\cite{wu2024unique3d} in~\cref{fig:ablation_diffusion}. As highlighted in the red boxes, \modelname (4-view) sometimes struggles with inconsistencies between the side views and the frontal view, leading to undesired artifacts such as breakages and non-watertight meshes. In contrast, our full 6-view model addresses these issues by incorporating additional intermediate views between the sides and the front. Meanwhile, Unique3D~\cite{wu2024unique3d} suffers from domain gaps, often producing unnatural geometries (e.g., flattened facial structures) and dull hair textures. 

\rebuttal{ \qheading{More Results of Multi-view Generation}
To further demonstrate the strong generalization ability of our six-view image and normal diffusion models across a wide range of styles, we present additional results of the generated six-view RGB and normal images in~\cref{fig:6view1}–\cref{fig:6view3}. }

\section{More Reconstruction Results}
We provide additional visual results of our method. \cref{fig:sup_hair} shows reconstruction results across a diverse range of hairstyles, while \cref{fig:sup_oil_painting} and \cref{fig:sup_style} present results on various artistic styles.

\input{supp/reconstruction}
\input{supp/sixview_images}

\end{appendix}

%% file: figs/data_gen.tex
\begin{figure}
    \centering
    \includegraphics[width=0.48\textwidth]{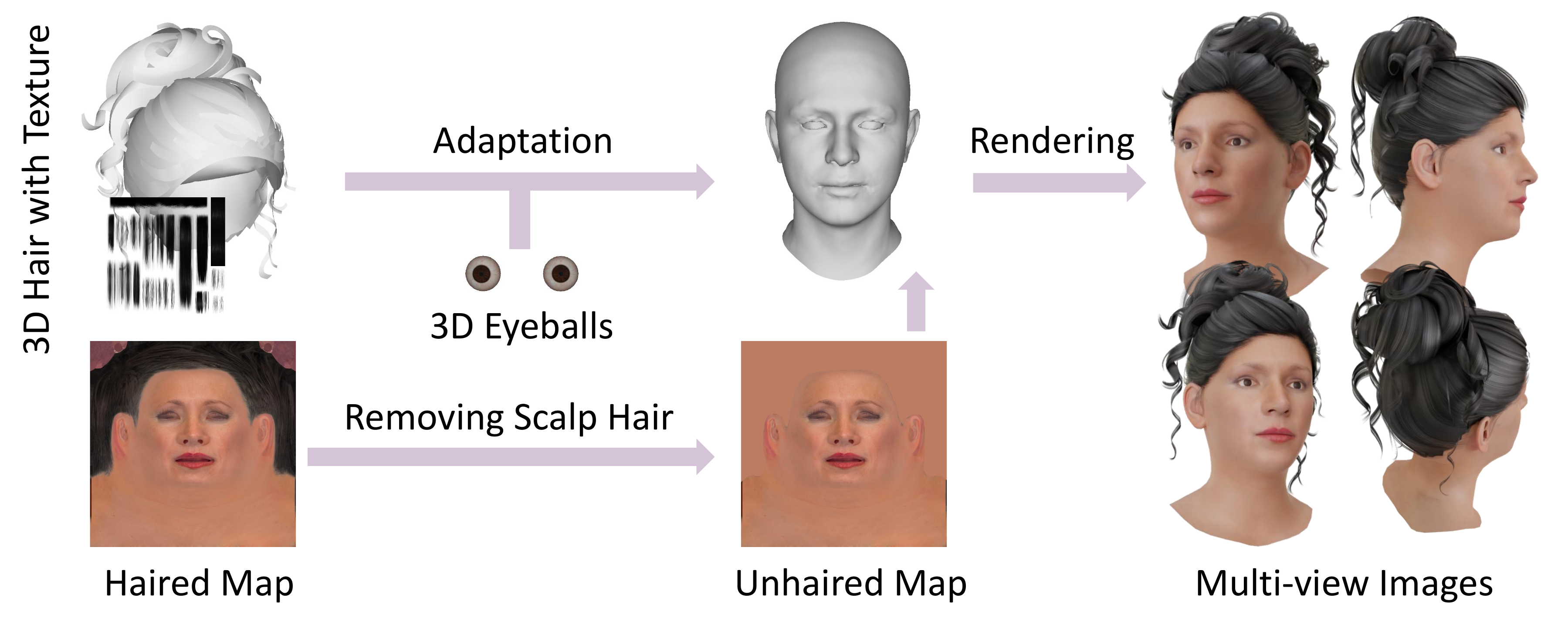}
    \caption{  
        Data generation pipeline for 3D head models with diverse hairstyles and facial features.   
    }
    \label{fig:data_gen}
\end{figure}

%% file: supp/data_example.tex
\begin{figure}
\centering
    \includegraphics[width=0.4\textwidth]{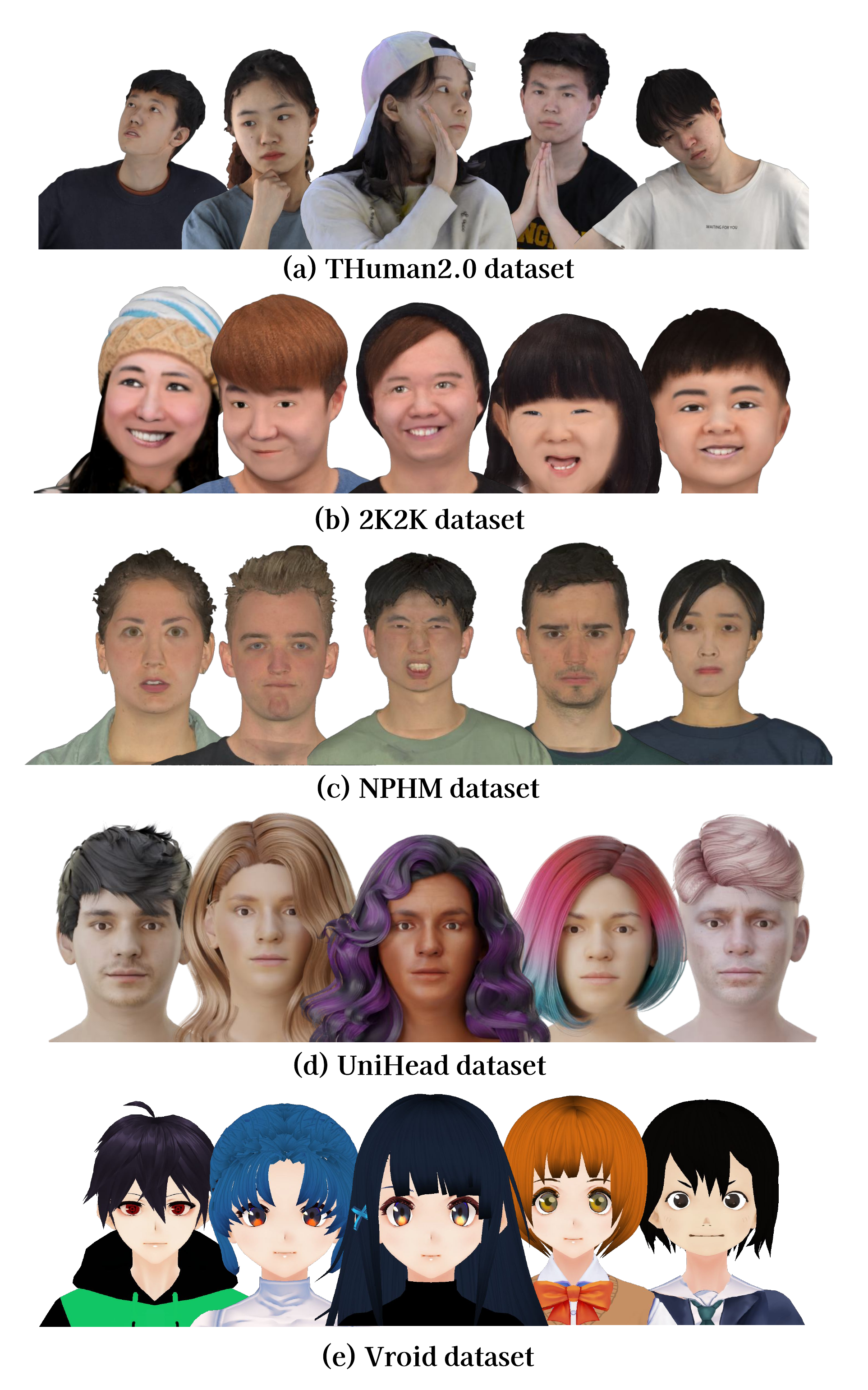}
    \caption{  
        Examples from different datasets.
    }
    \label{fig:data_example}
\end{figure}

%% file: figs/comparison_diffusion.tex
\begin{figure}
\centering
    \includegraphics[width=0.45\textwidth]{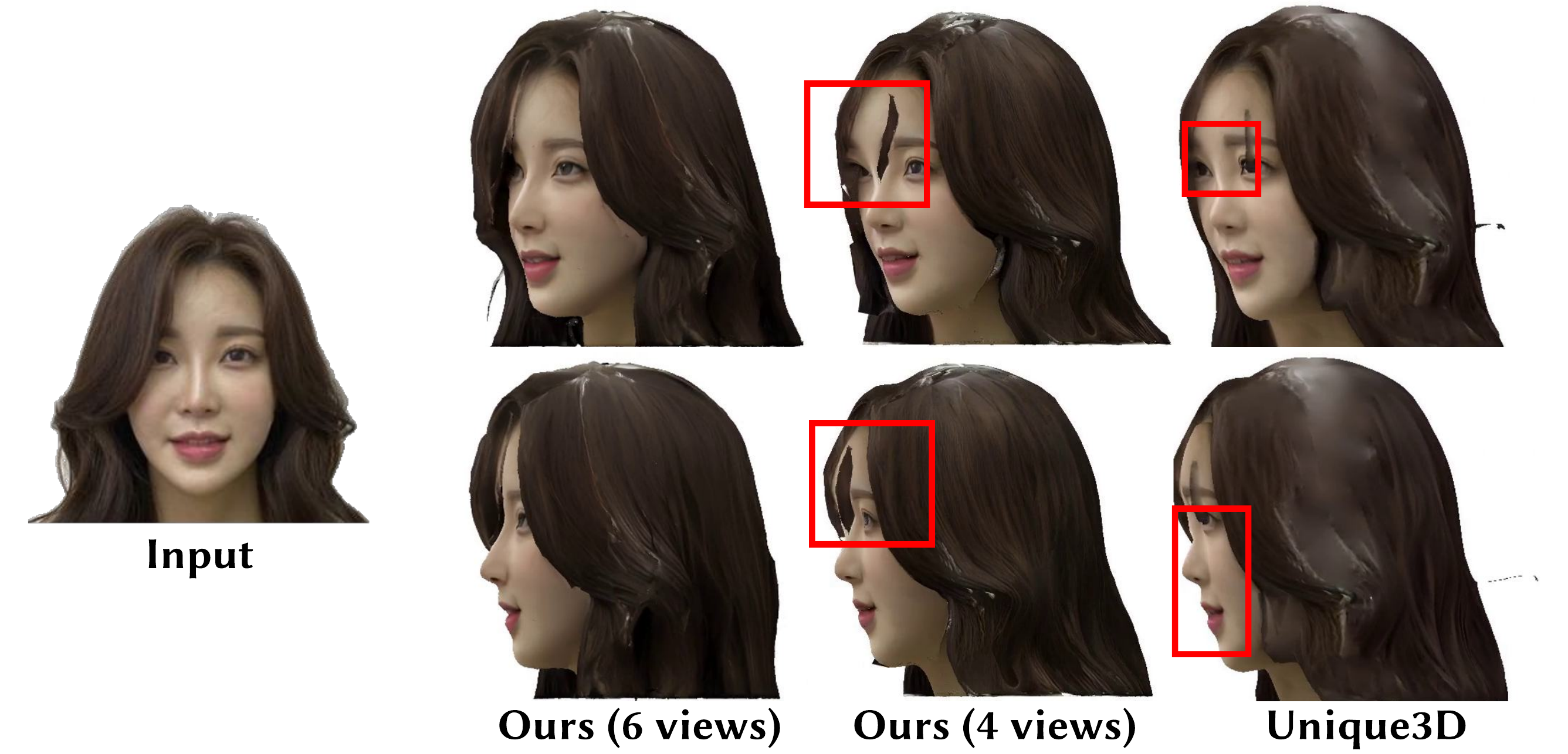}
    \caption{  
        Qualitative comparisons on diffusion models.  
    }
    \label{fig:ablation_diffusion}
\end{figure}

%% file: figs/comparison_unihair.tex
\begin{figure}
    \centering
    \includegraphics[width=0.48\textwidth]{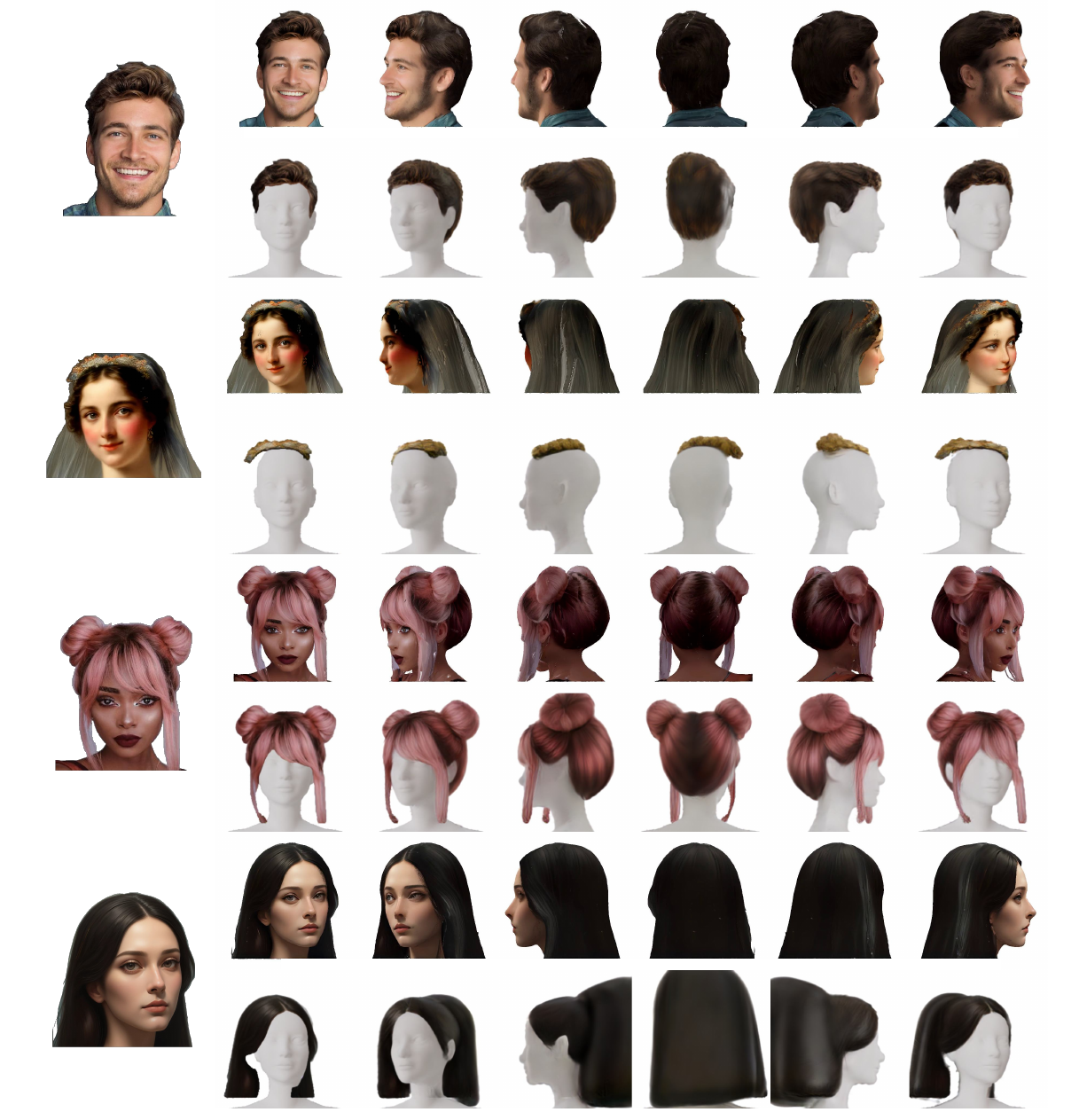}
    \caption{  
        \rebuttal{Qualitative comparisons with UniHair~\cite{zheng2024unihair}. In each example, the top row shows the results of our method and the bottom row shows the results of UniHair~\cite{wu2024unique3d}.}  
    }
    \label{fig:compare_unihair}
\end{figure}

%% file: figs/comparison_xportrait.tex
\begin{figure*}
    \centering
    \includegraphics[width=0.98\textwidth]{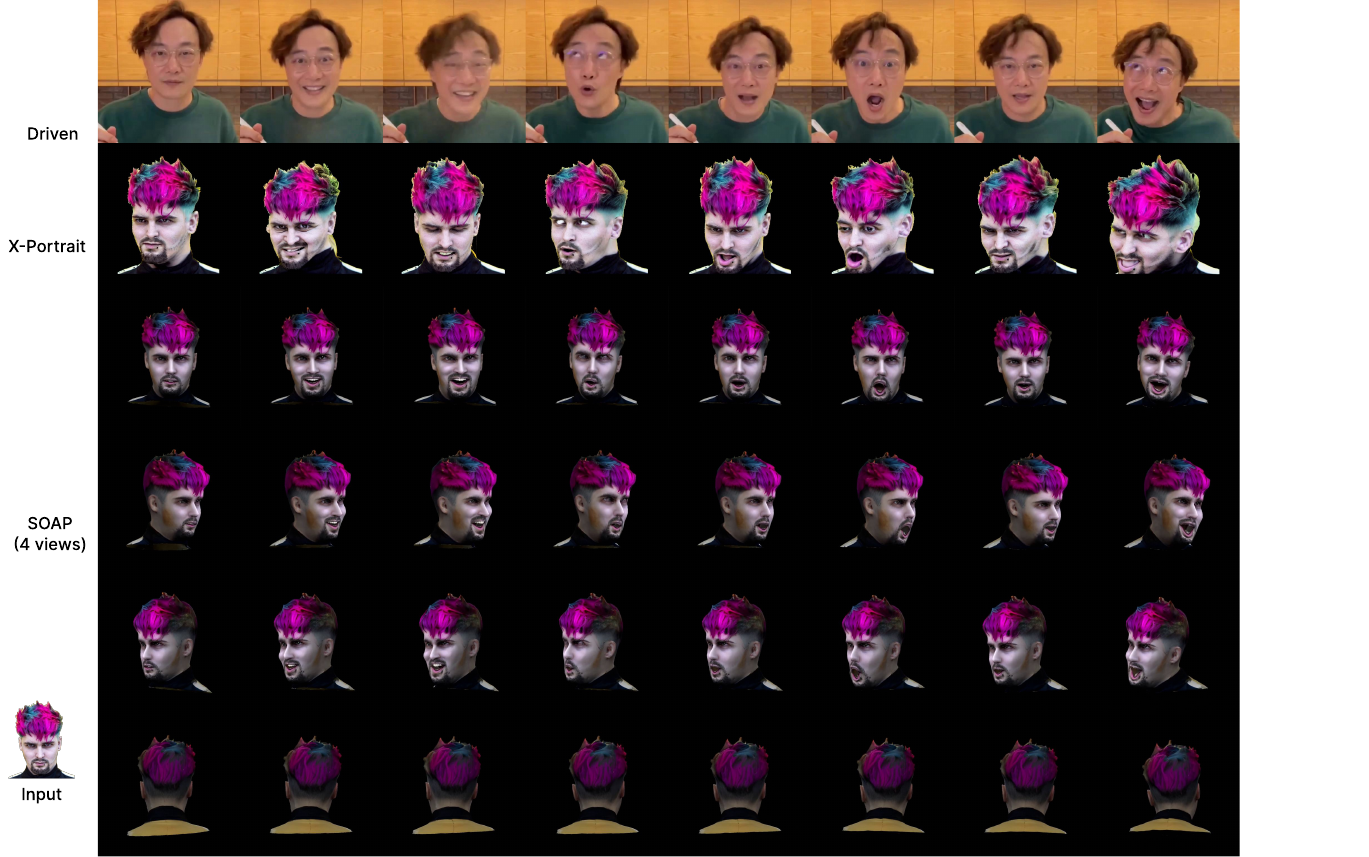}
    \caption{  
        \rebuttal{Qualitative comparisons with X-Portrait~\cite{xie2024x}.
        The first row displays random frames from the driving video, with the reference input image shown in the bottom-left corner.
        The second row presents the results of X-Portrait.
        The last four rows show the results of our method, rendered from different viewpoints.} 
    }
    \label{fig:compare_xportrait}
\end{figure*} 

%% file: figs/comparison_panic3d.tex
\begin{figure}
    \centering
    \includegraphics[width=0.48\textwidth]{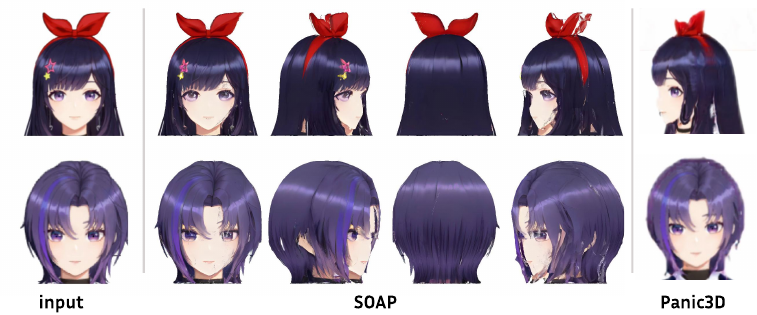}
    \caption{  
        \rebuttal{Qualitative comparisons with Panic3D~\cite{chen2023panic}.} 
    }
    \label{fig:compare_panic3d}
\end{figure}

%% file: figs/comparison_spherehead.tex
\begin{figure*}
    \centering
    \includegraphics[width=0.85\textwidth]{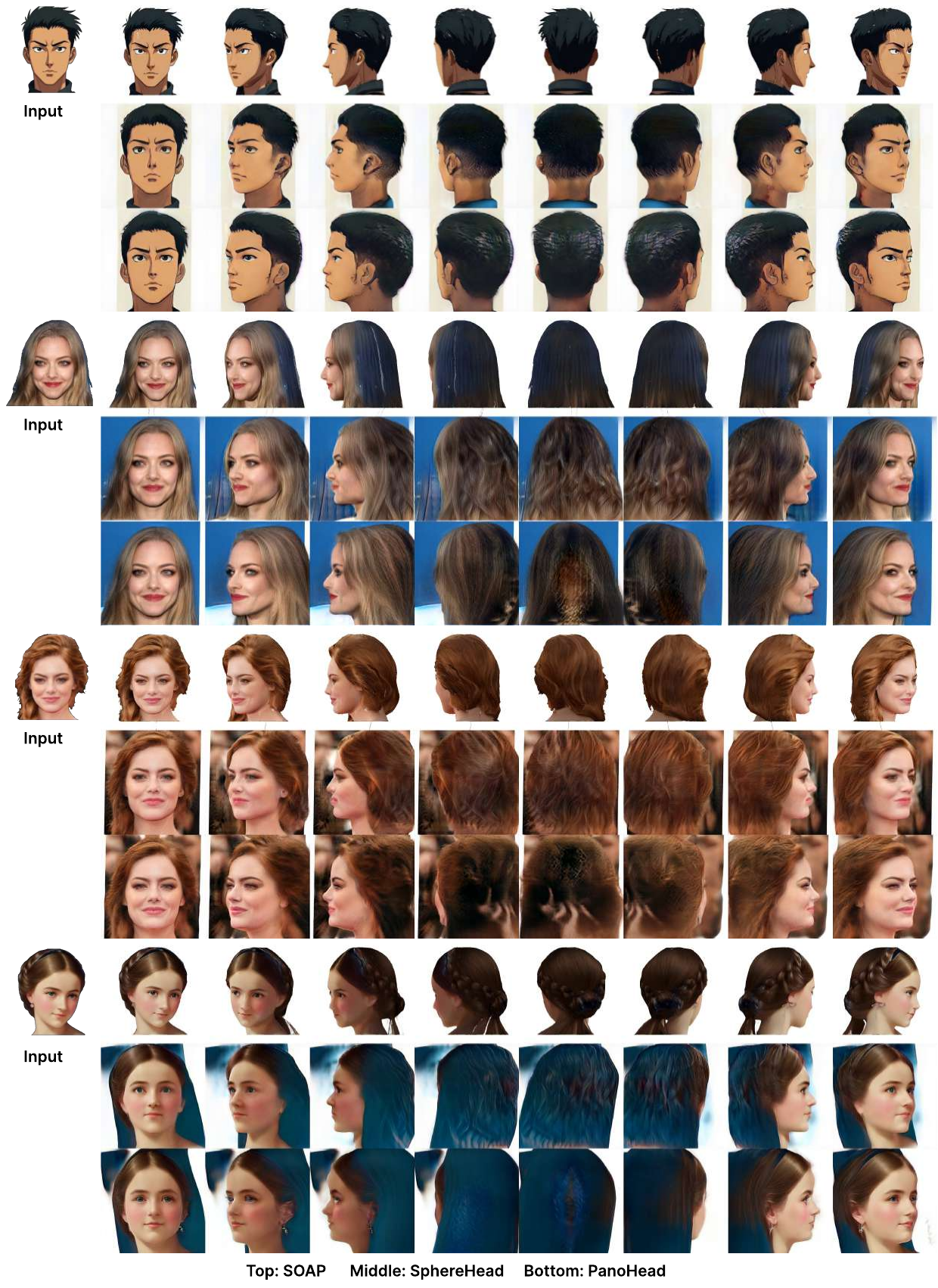}
    \caption{  
        \rebuttal{Qualitative comparisons with SphereHead~\cite{li2024spherehead} and PanoHead~\cite{An2023PanoHead}.}  
    }
    \label{fig:compare_gan}
\end{figure*} 

%% file: figs/rome_tab.tex
\begin{table}[htbp]
  \centering
  \caption{\rebuttal{Quantitative comparisons with ROME~\cite{Khakhulin2022ROME} across different views.} }
  \resizebox{\columnwidth}{!}{
  \begin{tabular}{l|ccc|ccc}
  
    \multirow{2}{*}{Azimuth} & \multicolumn{3}{c|}{Ours} & \multicolumn{3}{c}{ROME} \\
    & PSNR$\uparrow$ & SSIM$\uparrow$ & LPIPS$\downarrow$ & PSNR$\uparrow$ & SSIM$\uparrow$ & LPIPS$\downarrow$ \\
    \midrule
    $0^\circ$ & 21.77 & 0.8405 & 0.1144 & 16.07 & 0.7688 & 0.2344 \\
    $45^\circ$ & 16.27 & 0.7815 & 0.2161 & 13.95 & 0.7413 & 0.2819 \\
    $90^\circ$ & 17.10 & 0.7946 & 0.2015 & 12.16 & 0.7207 & 0.3511 \\
    $135^\circ$ & 15.73 & 0.7829 & 0.2332 & 10.64 & 0.6903 & 0.4058 \\
    $180^\circ$ & 19.62 & 0.8196 & 0.1570 & 9.78  & 0.6953 & 0.3675 \\
    $225^\circ$ & 15.82 & 0.7761 & 0.2355 & 12.16 & 0.7235 & 0.3435 \\
    $270^\circ$ & 17.24 & 0.7886 & 0.2009 & 12.24 & 0.7199 & 0.3594 \\
    $315^\circ$ & 16.67 & 0.7828 & 0.2155 & 13.82 & 0.7460 & 0.2888 \\
    \midrule
    $0^\circ$ (CSIM$\uparrow$) & \multicolumn{3}{c|}{0.8268} & \multicolumn{3}{c}{0.6202} \\

  \end{tabular}
  }
  \label{tab:rome}
\end{table}

%% file: figs/user_study_3d.tex
\begin{table}[h]
\centering
\scriptsize
\setlength{\tabcolsep}{3.5pt}
\renewcommand{\arraystretch}{1.2}
\caption{
\rebuttal{User study on 3D results of different methods. }
} 
\resizebox{0.97\linewidth}{!}{
\centering
\begin{tabular}{l|ccc}
Method                                            & View Consistency $\uparrow$  & ID Consistency $\uparrow$ & Overall Quality $\uparrow$ \\ \hline
ROME  & 1.668  & 1.978 & 1.714  \\  
SphereHead  & 2.956  & 2.890 & 2.822  \\  
Wonder3D  & 2.422  &  2.598 & 2.312  \\  
LGM  & 1.976  & 2.222 & 1.866  \\  
Unique3D  & 2.910  & 3.200 &  2.932  \\ \hline
\textbf{Ours}   & \textbf{4.756}       & \textbf{4.690}   & \textbf{4.714}    
\end{tabular}}
\vspace{-0.2 cm}
\label{table:user_study_3d}
\end{table}

%% file: figs/user_study_mv.tex
\begin{table}[h]
\centering
\scriptsize
\setlength{\tabcolsep}{3.5pt}
\renewcommand{\arraystretch}{1.2}
\caption{
\rebuttal{Quantitative evaluation of generated multi-view results of diffusion models with view/ID consistency (user study). }
} 
\resizebox{0.97\linewidth}{!}{
\centering
\begin{tabular}{l|ccc}
Method                                            & View Consistency $\uparrow$  & ID Consistency $\uparrow$ & Overall Quality $\uparrow$ \\ \hline
Unique3D  & 3.154  & 3.404 & 3.178  \\ \hline
\textbf{Ours}   & \textbf{4.801}  & \textbf{4.825}   & \textbf{4.854}    
\end{tabular}}
\vspace{-0.2 cm}
\label{table:user_study_mv}
\end{table}

%% file: supp/reconstruction.tex
\begin{figure*}
\centering
    \includegraphics[width=\textwidth]{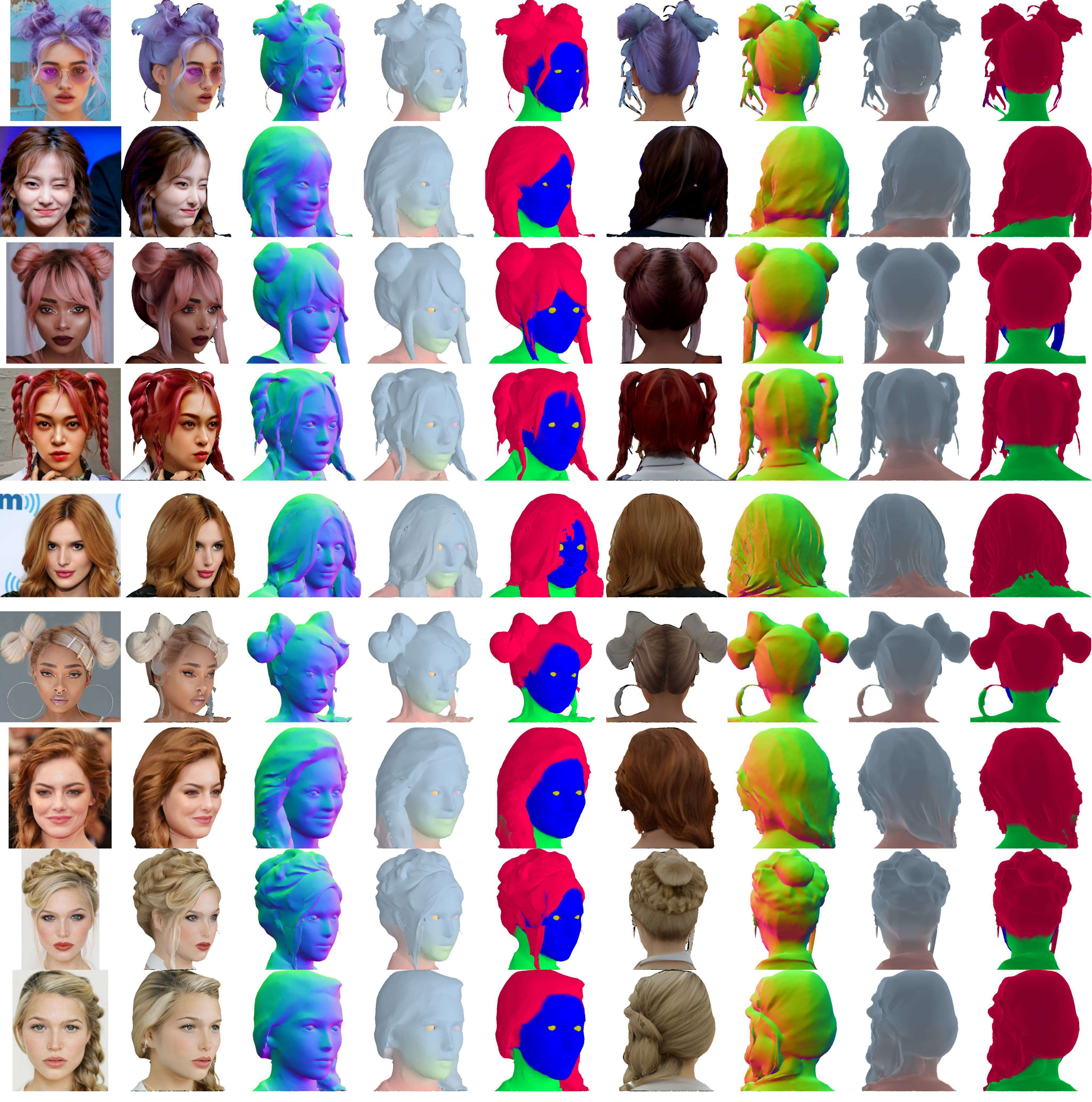}
    \caption{  
        More reconstruction results. From left to right are the input image, the rendered RGB, normal, skinning weights, and parsing labels. 
    }
    \label{fig:sup_hair}
\end{figure*}

\begin{figure*}
\centering
    \includegraphics[width=\textwidth]{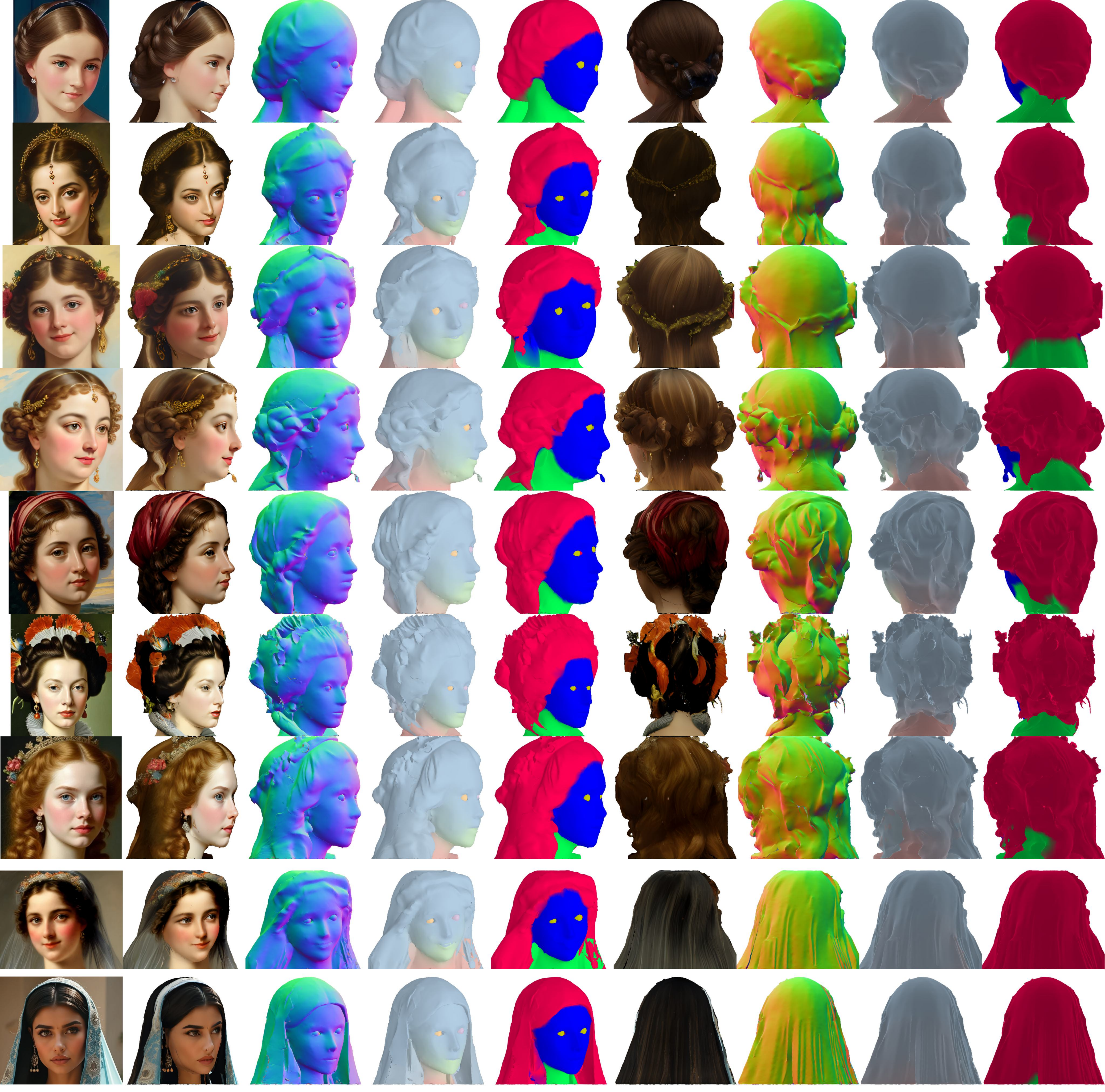}
    \caption{  
        More reconstruction results. From left to right are the input image, the rendered RGB, normal, skinning weights, and parsing labels. 
    }
    \label{fig:sup_oil_painting}
\end{figure*}

\begin{figure*}
\centering
    \includegraphics[width=\textwidth]{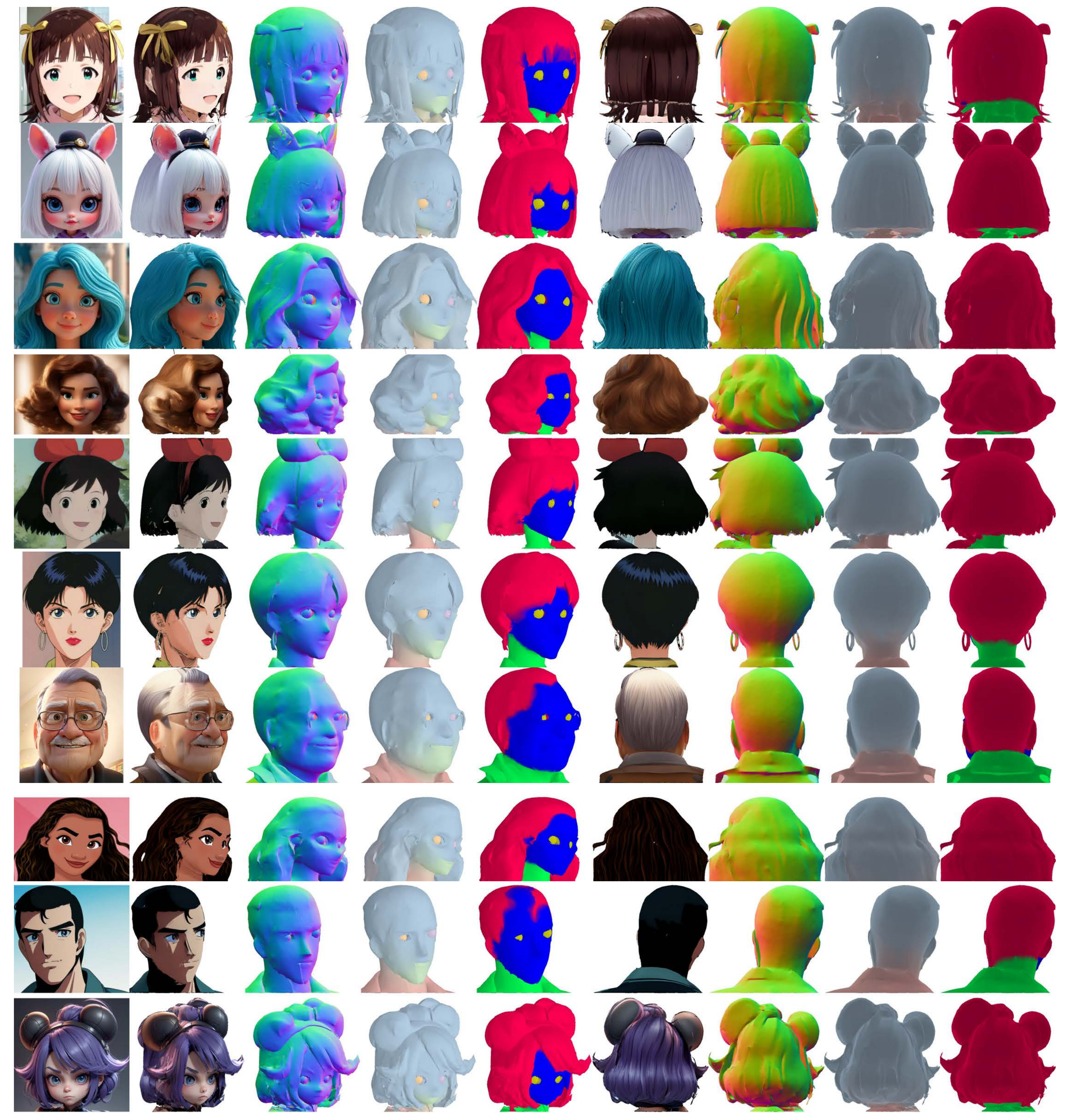}
    \caption{  
        More reconstruction results. From left to right: the input image, rendered RGB, normals, skinning weights, and parsing labels.
    }
    \label{fig:sup_style}
\end{figure*}

%% file: supp/sixview_images.tex
\begin{figure*}
\centering
    \includegraphics[width=\textwidth]{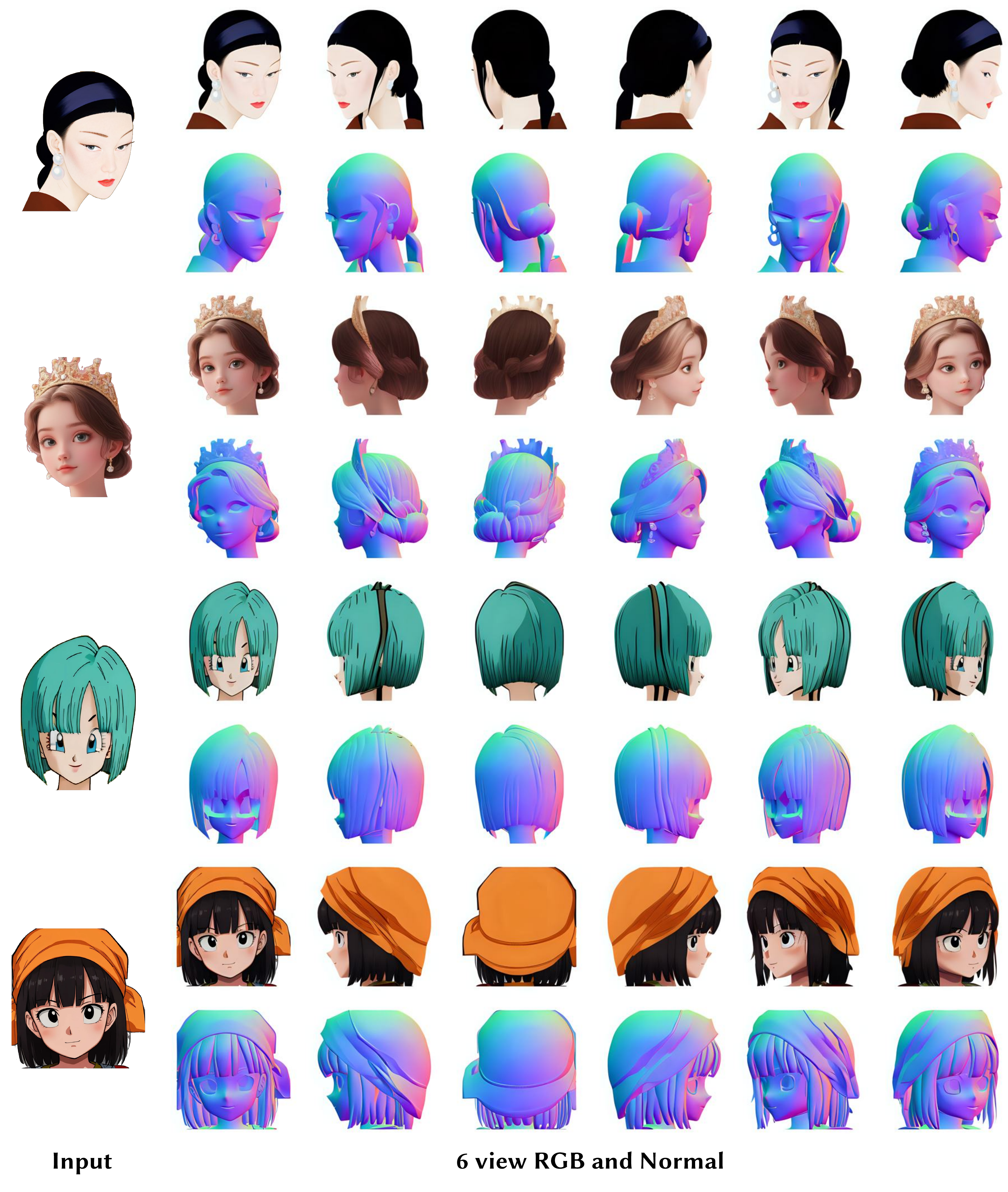}
    \caption{  
        More results of six-view RGB images and normal maps generated by our diffusion model.  
    }
    \label{fig:6view1}
\end{figure*}
\begin{figure*}
\centering
    \includegraphics[width=\textwidth]{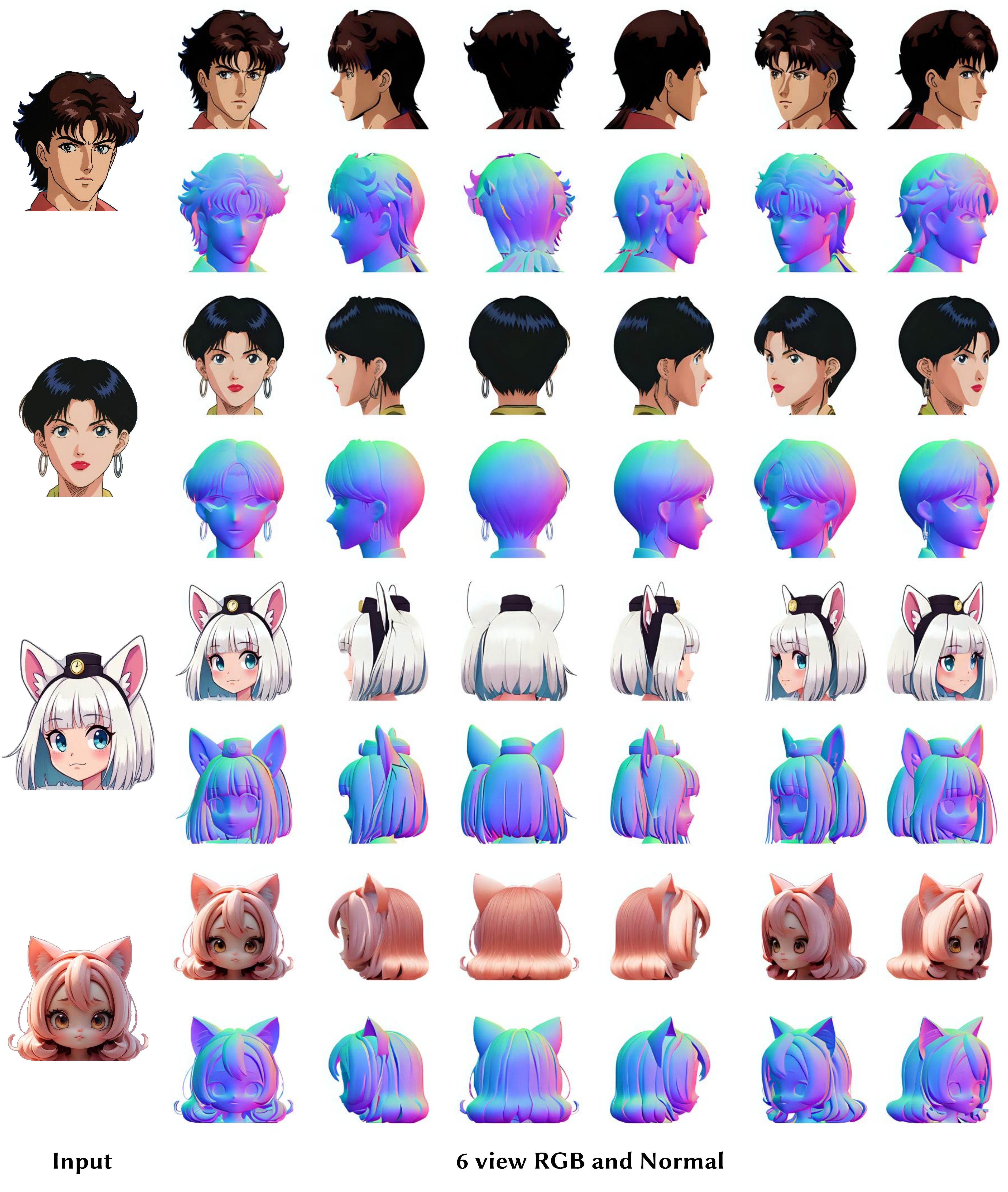}
    \caption{  
        More results of six-view RGB images and normal maps generated by our diffusion model.  
    }
    \label{fig:6view2}
\end{figure*}
\begin{figure*}
\centering
    \includegraphics[width=\textwidth]{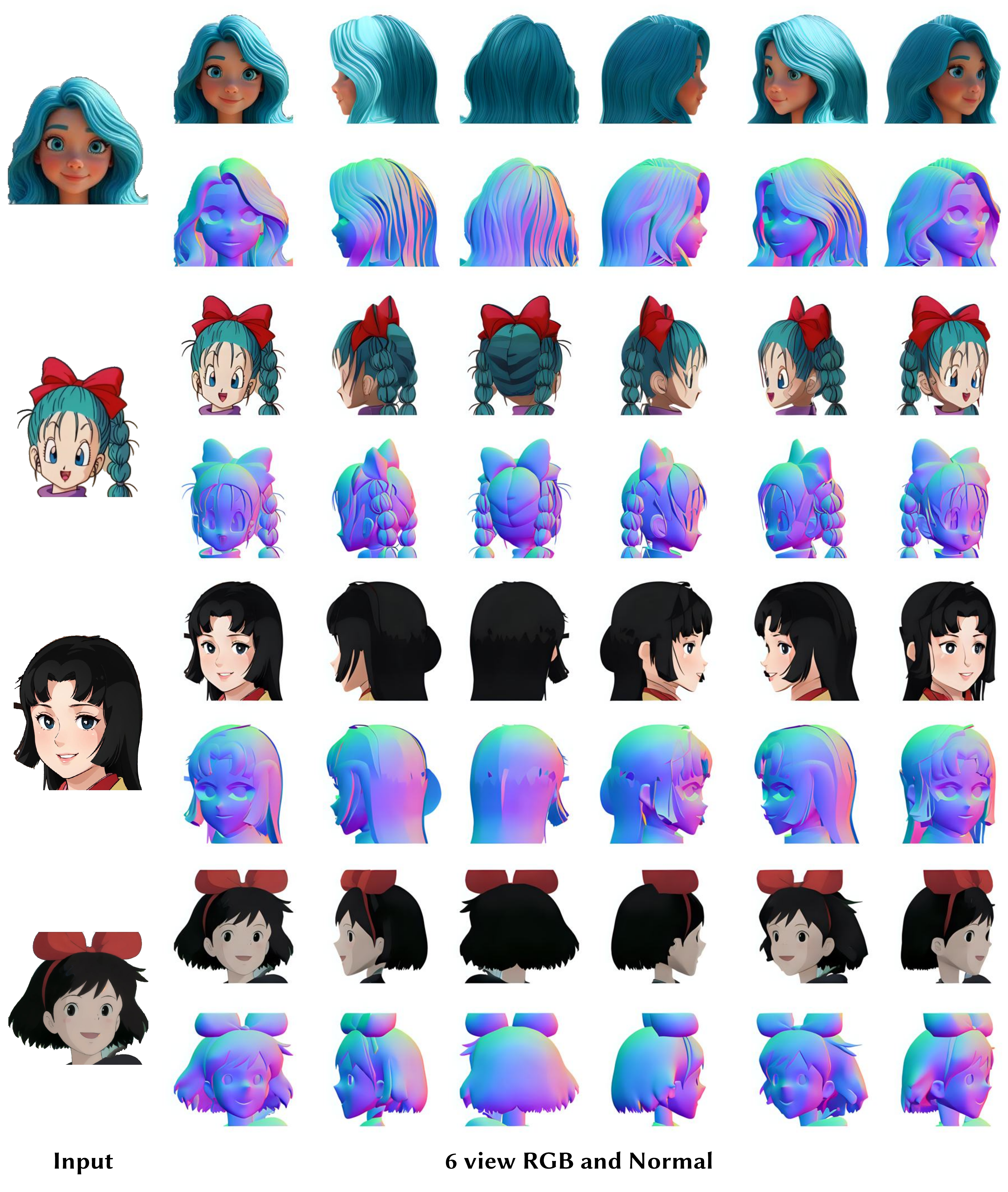}
    \caption{  
        More results of six-view RGB images and normal maps generated by our diffusion model.  
    }
    \label{fig:6view3}
\end{figure*}